\pdfoutput=1

\documentclass[11pt]{article}

\usepackage[preprint]{acl}

\usepackage{times}
\usepackage{latexsym}
\usepackage{amsmath}
\usepackage[T1]{fontenc}

\usepackage[utf8]{inputenc}

\usepackage{microtype}
\usepackage{tcolorbox}
\usepackage{colortbl} 
\usepackage{newfloat}
\usepackage{listings}
\usepackage{inconsolata}

\usepackage{graphicx}
\usepackage{booktabs}

\title{FamilyTool: A Multi-hop Personalized Tool Use Benchmark}

\author{
 \textbf{Yuxin Wang*\textsuperscript{1,2}},
 \textbf{Yiran Guo*\textsuperscript{1}},
 \textbf{Yining Zheng\textsuperscript{1}},
 \textbf{Zhangyue Yin\textsuperscript{1}},
\\
 \textbf{Shuo Chen\textsuperscript{1}},
 \textbf{Jie Yang\textsuperscript{1}},
 \textbf{Jiajun Chen\textsuperscript{1}},
  \textbf{Yuan Li\textsuperscript{1}},
 \textbf{Xuanjing Huang\textsuperscript{1,2}},
 \textbf{Xipeng Qiu\textsuperscript{1}}
 \\
 \textsuperscript{1}Computer Science, Fudan University,\\
 \textsuperscript{2}Institute of Modern Languages and Linguistics, Fudan University\\
\\
 \small{
  * Equal Contribution
 }
}

\begin{document}
\maketitle
\begin{abstract}
The integration of tool learning with Large Language Models (LLMs) has expanded their capabilities in handling complex tasks by leveraging external tools. However, existing benchmarks for tool learning inadequately address critical real-world personalized scenarios, particularly those requiring multi-hop reasoning and inductive knowledge adaptation in dynamic environments. To bridge this gap, we introduce FamilyTool, a novel benchmark grounded in a family-based knowledge graph (KG) that simulates personalized, multi-hop tool use scenarios. FamilyTool, including base and extended datasets, challenges LLMs with queries spanning from 1 to 4 relational hops (e.g., inferring familial connections and preferences) and 2 to 6 hops respectively, and incorporates an inductive KG setting where models must adapt to unseen user preferences and relationships without re-training, a common limitation in prior approaches that compromises generalization. We further propose KGETool: a simple KG-augmented evaluation pipeline to systematically assess LLMs' tool use ability in these settings. Experiments reveal significant performance gaps in state-of-the-art LLMs, with accuracy dropping sharply as hop complexity increases and inductive scenarios exposing severe generalization deficits. These findings underscore the limitations of current LLMs in handling personalized, evolving real-world contexts and highlight the urgent need for advancements in tool-learning frameworks. FamilyTool serves as a critical resource for evaluating and advancing LLM agents' reasoning, adaptability, and scalability in complex, dynamic environments. Code and dataset are available at \href{https://github.com/yxzwang/FamilyTool}{https://github.com/yxzwang/FamilyTool}.
\end{abstract}

\section{Introduction}

Large Language Models (LLMs) have emerged as powerful tools in the field of artificial intelligence, particularly in natural language processing (NLP). Tool use, is obviously a large and suitable domain for LLMs, involving equipping LLMs with the ability to interact with external tools to perform tasks more effectively. This integration of tool learning aims to address limitations in traditional LLMs, such as their ability to handle complex reasoning or access real-time information. 

\begin{figure*}[htbp]
    \centering
    \includegraphics[width=\linewidth]{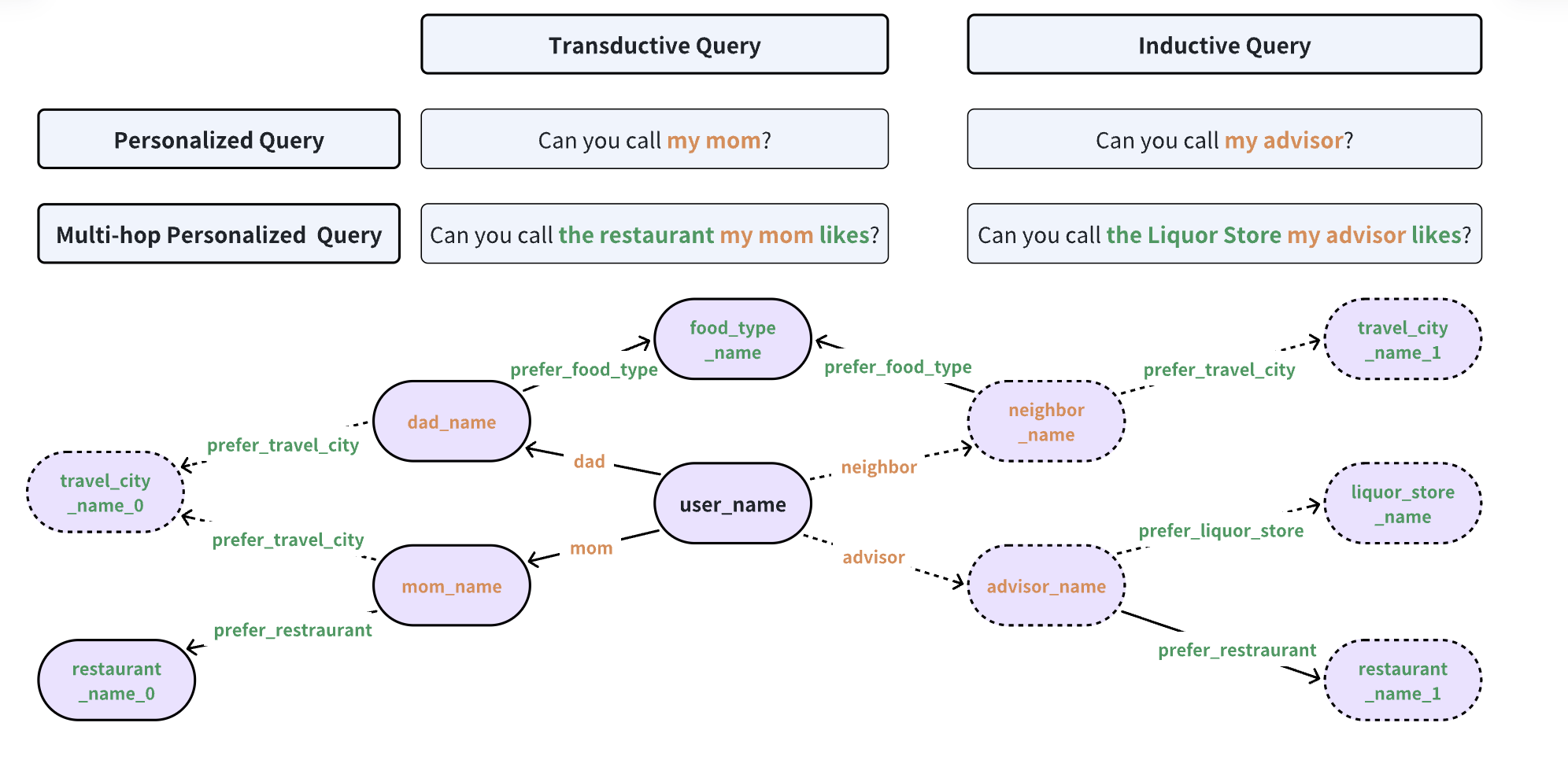}
    \caption{Multi-hop personalized inductive queries for family-specific tool use. The dashed edges in the knowledge graph represent inductive relations. }
    \label{fig:multihop_psersonalized_inductive}
\end{figure*}

To evaluate LLMs' tool use ability, a lot of benchmarks~\cite{schick_toolformer_2023,liu_toolace_2024} have been proposed considering different parts of tool use, including different scenarios ranging from scientific fields~\cite{ma_sciagent_2024} to device-side tool use~\cite{wang_appbench_2024_1,lu_toolsandbox_2024,shen_shortcutsbench_2024_1}. In device-side tool use scenerio, \textit{family-specific} tool use is always needed in commercial use to make it convenient for the user. However, there is no benchmark covering this scenario, which features two important properties from our side:  \textbf{Multi-hop Reasoning} and \textbf{Inductive Reasoning}. We show the example of multi-hop and inductive queries in Figure \ref{fig:multihop_psersonalized_inductive} with detailed demonstration below.

\textbf{Multi-hop Personalized Reasoning.} Multi-hop Reasoning comes from the relationship in the family members and other inter-person relationships, as well as human preferences and is used in tool use parameter generation. For example as shown in Figure \ref{fig:multihop_psersonalized_inductive}, when the user wants to make a phone call, he may want to call his mom, or his mom's friend, or his mom's favourate restaurant for dinner booking. In last two quests, we need a multi-hop reasoning before tool use. To achieve multi-hop reasoning, we use \textbf{Knowledge Graph} (KG) to store the information of user family, which we call family KG.

\textbf{Inductive Reasoning.} The family-specific tool use scenario will inevitably meet the updating of user's behavior like making new preferences and adding new relationships that are totally unknown before, which is similar to the inductive setting of existing KG tasks like link prediction~\cite{DBLP:conf/icml/TeruDH20}. For example in Figure \ref{fig:multihop_psersonalized_inductive}, the dashed lines and nodes represent unknown relations and entities respectively in the training. Previous works~\cite{li_api-bank_2023_1, tang_toolalpaca_2023, qin_toolllm_2023} in tool use may handle unknown tools with documents, but hard to handle the situation for unknown relations in KG with training. It is very time-consuming if we train a new LLM for every updating of KG. So for generalization, we emphasize the inductive reasoning.

In the end, we propose \textbf{FamilyTool}: A Multi-hop Personalized Tool Use Benchmark. Our FamilyTool includes base dataset \textbf{FamilyTool-e} and extended one \textbf{FamilyTool-e}, ranging the hop number required from 1 to 6 in the queries for tool use. Also, we propose a simple KG-augmented LLM Tool use pipeline: \textbf{KGETool}, for the totally inductive setting for KG and evaluate common LLMs using KGETool. Experiments show great difficulty for existing LLMs and arouse a great challenge to LLM agents.

We organize our paper in the following way. First we show the generation pipeline of FamilyTool, including data examination, Then we demonstrate KGETool pipeline for evaluating LLMs. Last, we analyze the experiments which show that existing LLMs are not strong enough for such problems and urge for new better LLMs. 
\section{Related Works}
\textbf{In-context Tool Use.} In-context tool use~\cite{YiningZHENG:0} requires the model to rely on tool-related information provided in the context during reasoning, intrinsically including the concept of Retrieval Augmented Generation (RAG) for LLMs. Research on in-context tool use is mainly divided into two categories: training-free works \cite{gao_pal_2022_1, shen_hugginggpt_2023_1, yang_gpt4tools_2023_1} leverages the strong instruction-following capabilities of models like ChatGPT by providing candidate tool information during the inference phase to achieve dynamic tool invocation; post-training works \cite{li_api-bank_2023_1, tang_toolalpaca_2023, qin_toolllm_2023, dubey_llama_2024_1} employs generalization training to enable the model to generate tool use instructions based on tool information provided in the context. This generalization training approach allows the model to generate tool call it has never encountered during training, significantly enhancing the flexibility and adaptability of its tool usage. The simple pipeline in our paper falls into the first category. 

Following Toolformer~\cite{schick_toolformer_2023}, a lot of tool use benchmarks emerged, mainly focusing on general tool use and tool generation~\cite{song_restgpt_2023,qin_toolllm_2023,huang_metatool_2024_1,patil_gorilla_2023,xu_tool_2023,tang_toolalpaca_2023,chen_t-eval_2024_1,shen_taskbench_2023_1,basu_api-blend_2024_1,li_api-bank_2023_1,ye_tooleyes_2025_1}, tool-tool relationship~\cite{huang_planning_2024_1,wu_seal-tools_2024_1,liu_toolace_2024,wang_mtu-bench_2024}, robustness~\cite{wang_learning_2024,ye_toolsword_2024_1,ye_rotbench_2024_1,guo_stabletoolbench_2024}, and specific scenarios like ambiguous user queries~\cite{farn_tooltalk_2023}, scientific questions~\cite{ma_sciagent_2024}, multi-modal queries~\cite{wang_mllm-tool_2024} and phone-use queries~\cite{wang_appbench_2024_1,lu_toolsandbox_2024,shen_shortcutsbench_2024_1}. Recently a new paper~\cite{hao2025evaluatingpersonalizedtoolaugmentedllms} proposed a personalized tool use benchmark. Although they have different user files, their tool use only focuses on a single user, not a family of users. All in all, there is no existing benchmark focusing on the multi-hop personalized inductive setting of tool use parameters, which is the main contribution of our paper.

\textbf{Multi-hop Question Answering.}
Multi-hop Question Answering (QA) is a well-known task in NLP domain, with a lot of datasets including HotpotQA~\cite{yang2018hotpotqa}, 2WikMultihopQA (2WMQA)~\cite{ho-etal-2020-constructing}, CAG~\cite{pan2024not}, PopQA~\cite{mallen2022not},  WebQuestions~\cite{berant2013semantic}, MuSiQue~\cite{trivedi2022musique} and LV-Eval~\cite{yuan2024lv}. They mainly focus on retrieving gold documents from a bunch of documents in each hop for question answering. If we view Knowledge Graph Question Answering (KGQA) as a type of Multi-hop QA with Knowledge Graph as database. Related datasets include WebQSP~\cite{yih2016value}, CWQ~\cite{talmor2018web} and MetaQA~\cite{zhang2018variational}. Thought those KGs are from real world, they are far from personalized scenarios, let alone they have human-made meaningless relation types with no application value.

\textbf{Inductive KG Datasets.} 
 For a long time, inductive KG methods have focused on unknown nodes and edges (not relations) from Grail \cite{DBLP:conf/icml/TeruDH20} creating inductive version of WN18RR \cite{dettmers2018conve}, FB15k-237 \cite{toutanova2015representing}, and NELL-995 \cite{deeppath}. Mainly two types of methods are proposed, including rule-based methods~\cite{DBLP:conf/ijcai/MeilickeCRS19, DBLP:journals/vldb/GalarragaTHS15,DBLP:conf/ijcai/OmranWW18,DBLP:conf/aaai/Pirro20,DBLP:conf/nips/YangYC17,DBLP:conf/nips/SadeghianADW19,DBLP:conf/acl/NeelakantanRM15,DBLP:conf/eacl/McCallumNDB17,DBLP:conf/iclr/DasDZVDKSM18,DBLP:conf/iclr/QuCXBT21} and GNN-based methods \cite{DBLP:conf/icml/TeruDH20,DBLP:conf/aaai/MaiZY021,DBLP:conf/nips/ZhuZXT21,DBLP:conf/nips/LiuGHK21,DBLP:conf/icml/YanMGT022,DBLP:conf/www/ZhangY22,DBLP:conf/iclr/0001DWH22}. \cite{pmlr-v202-lee23c} and \cite{geng2023relational} generates fully inductive KG datasets based on existing transductive ones for examining their methods. \cite{cui2023lifelong} proposed a life-long learning dataset for KG link prediction including unknown relations.  There is no inductive KG datasets for tool use for now.

\section{Preliminaries}

\subsection{Multi-hop Reasoning}
Multi-hop Question Answering~\cite{yang2018hotpotqa} is an NLP task with history. It requires the resources from different documents with a specific order to generate the right answer. In LLM era, reasoning is a better and larger word for multi-hop problems, including more than answering questions. Unlike single-step reasoning, which retrieves answers directly from a single piece of evidence, multi-hop reasoning requires navigating through interconnected facts, inferences, or contexts to answer a question. This capability is critical for addressing complex real-world problems where answers depend on implicit relationships or distributed knowledge.

In Tool use domain, multi-hop reasoning usually appears at serial tool use. However, there is another multi-hop scenario that is neglected, as which we show in Figure \ref{fig:multihop_psersonalized_inductive} in the second row. For instance, tool call with a query like, "Can you call the restaurant my mom likes?" demands first get the mother of user and then search which restaurant she prefers. This is the scenario for one tool call with multi-hop reasoning parameters. One may argue that this can be solved by multi-round tool use, but definitely the user will prefer one direct tool use than talking with the agent further to help it find the parameter. So multi-hop reasoning for the tool use parameter is important in real world scenarios.

\subsection{Transductive and Inductive KG reasoning}
Another feature we focus on is the inductive KG reasoning, which is totally different with transductive one. They differ in their assumptions about the entities and relations encountered during training and inference, shaping their applicability to real-world scenarios.

\textbf{Transductive Reasoning.} Transductive reasoning in KGs operates under a closed-world assumption, where all entities and relations are known during training. In traditional tasks, it focuses on predicting missing links (e.g., completing triples like (head, relation, tail) where tail has a candidate set) within a fixed graph structure using embedding-based methods such as TransE~\cite{DBLP:conf/nips/BordesUGWY13}, RotatE~\cite{DBLP:conf/iclr/SunDNT19}, or ComplEx~\cite{TrouillonWRGB_2016} or Graph Neural Networks~\cite{DBLP:conf/esws/SchlichtkrullKB18}. These methods encode entities and relations into low-dimensional vectors to capture semantic patterns, enabling tasks like link prediction or triple classification. While effective for static KGs with predefined entities and relations, transductive approaches cannot generalize to unseen entities or relations, limiting their applicability in dynamic environments where new data frequently emerges, which is exactly the property of personalized scenario we focus on.

\textbf{Inductive Reasoning.} Opposite to transductive setting, inductive reasoning addresses the open-world challenge by generalizing to unseen entities and relations not present during training~\cite{DBLP:conf/icml/TeruDH20}. It leverages structural patterns and rule induction~\cite{DBLP:conf/eacl/McCallumNDB17,DBLP:conf/iclr/DasDZVDKSM18,DBLP:conf/iclr/QuCXBT21}, or well-designed GNNs~\cite{DBLP:conf/aaai/MaiZY021,DBLP:conf/nips/ZhuZXT21} to infer relationships in evolving or partially observed KGs. As we shown in Figure \ref{fig:multihop_psersonalized_inductive} right column, the query demands the usage of inductive KGs with dashed lines in the figure, which is very common when the user adds a new friend or meets a new person, or finds a new interest. Inductive KG reasoning has spent years~\cite{DBLP:conf/nips/LiuGHK21,DBLP:conf/icml/YanMGT022,DBLP:conf/iclr/0001DWH22} focusing on datasets with only unknown entities and not until recently unknown relations come into the field~\cite{pmlr-v202-lee23c,geng2023relational}, raising new challenges to inductive reasoning in KGs. Luckily in our scenario we apply LLMs with very good generalization ability for inductive reasoning.

\section{FamilyTool Generation}
In this section, we demonstrate in detail how we construct FamilyTool, which is splited in the pipeline of KG generation, Query-Answer Pair Generation and Data Examination.
\subsection{KG Generation}
First, we need to get the family-specific knowledge graph for FamilyTool. For a simple start, we first generate a family of three persons: Bob, Jack and Alice, where Jack and Alice are couple and Bob is their son. Then, we need to generate the preferences of these three persons.

\textbf{Preference Generation.} We extract preferences from an existing tool use Benchmark: MTUBench~\cite{wang_mtu-bench_2024}, from which we select our tool. We first use GPT-4o to extract candidate preference set $\mathcal{P}_{candidate}$, given a selected API document. Then we filter some unreasonable and duplicated preferences and split the preferences in several categories to make small KGs. The entities of preferences are signals for easy use. For example, Bob's preferred food is food$\_$0000.

\subsection{Query-Answer Pair Generation}
With the small KGs and selected tool use document, we can generate the Query-Answer Pair with GPT-4o. We let GPT-4o to use the KG as much as possible in query generating and assign the speaker for each query. And then we give it the query and make it to generate the answer in json format.

\subsection{Data Examination} 
We use three steps for data examination. First we filter the datas with hallucination in the KGs. Then we use GPT-4o to check data from several directions, including the examination of query and answer respectively. Quality check of the query determines if it is a valid request command. The quality inspection of the target involves a comprehensive evaluation encompassing the following criteria:  
\begin{itemize}
    \item Syntax check  assessing whether the call format adheres to the prescribed structural and syntactical requirements.  
    \item  Compliance check with query intent determining if the tool call accurately reflects the intended purpose and semantic meaning of the original query.  
    \item Parameter standardization check verifying that all supplied parameters conform to predefined specifications in terms of data type, range, and mandatory/optional constraints for the tool use document.  
    \item Feasibility check evaluating whether the user's request falls within the functional and operational boundaries of the tool's defined capabilities.  
\end{itemize}

This systematic examination ensures robustness, correctness, and operational feasibility in tool use data generation. The last step we use \textbf{human check} to guarantee the goodness of samples. In the end, we get the base dataset for FamilyTool, denoted as \textbf{FamilyTool-b}. Then we extract the used links in all samples to get the accompanied KG, which is denoted as $\mathcal{G}_b$.

\subsection{Data Extension}

We extend the $\mathcal{G}_b$ to fit more complicated real-world scenerio. We add five new people including Jack and Alice's parents and Bob's teacher and their preferences and denote the new KG as $\mathcal{G}_e$. With $\mathcal{G}_e$ and FamilyTool-b, we use GPT-4o to extend the queries and use the new links in $\mathcal{G}_e$.After filtering datas with hallucination and human check, we get this extended dataset, denoted as \textbf{FamilyTool-e}.  

\subsection{Statistics}
We show the statistics of FamilyTool-b and FamilyTool-e and their accompanied family KG in Table \ref{tab:statistics-FamilyTool}. The tool documents are from MTU-Bench~\cite{wang_mtu-bench_2024}, while some tools have the same name but with different parameters, we just choose one. 
\begin{table}[htbp]

    \centering
    \resizebox{0.45\textwidth}{!}{
        \begin{tabular}{l|cc|ccc}
            \toprule
           & \multicolumn{2}{c}{Tool Statistics}& \multicolumn{3}{c}{KG Statistics}\\
            \midrule
            &Tool Number & Test Samples & Node. & Edge. & Edge Types   \\
            \midrule
            FamilyTool-b &102 & 483 & 154  & 147 & 65  \\
            FamilyTool-e &102 & 455 & 491  & 480 & 66  \\
            \bottomrule
        \end{tabular}
    }
    \caption{FamilyTool and family KG statistics.}
    \label{tab:statistics-FamilyTool}
\end{table}

Then we show the hops of two datasets in Table \ref{tab:statistics-hopdistribution}. In FamilyTool-b, for one parameter that needs KG knowledge, the hop number is at least one (for relationship), which is two in FamilyTool-e because we enforce two hops for relationships. In addition, we intend to introduce questions that needs two paramters that demand KG knowledges, such as "Can you book a ticket from my mother's living city to her favourite city?", demanding two parameters at one tool call from multi-hop reasoning.

\begin{table}[htbp]

    \centering
    \resizebox{0.45\textwidth}{!}{
    \begin{tabular}{l|cccccc}
        \toprule
         Hop Number & 1 & 2& 3 &4&5&6  \\
        \midrule
        FamilyTool-b & 106 &361  & 13 &3&0&0 \\
        FamilyTool-e & 0 &36  & 387&22&7&3  \\
        \bottomrule
    \end{tabular}}
    \caption{Hop distribution of FamilyTool.}
    \label{tab:statistics-hopdistribution}
\end{table}

\section{KGETool Pipeline}
\label{sec:simplepipeline}

\begin{figure}[t]
    \centering
    \includegraphics[width=\linewidth]{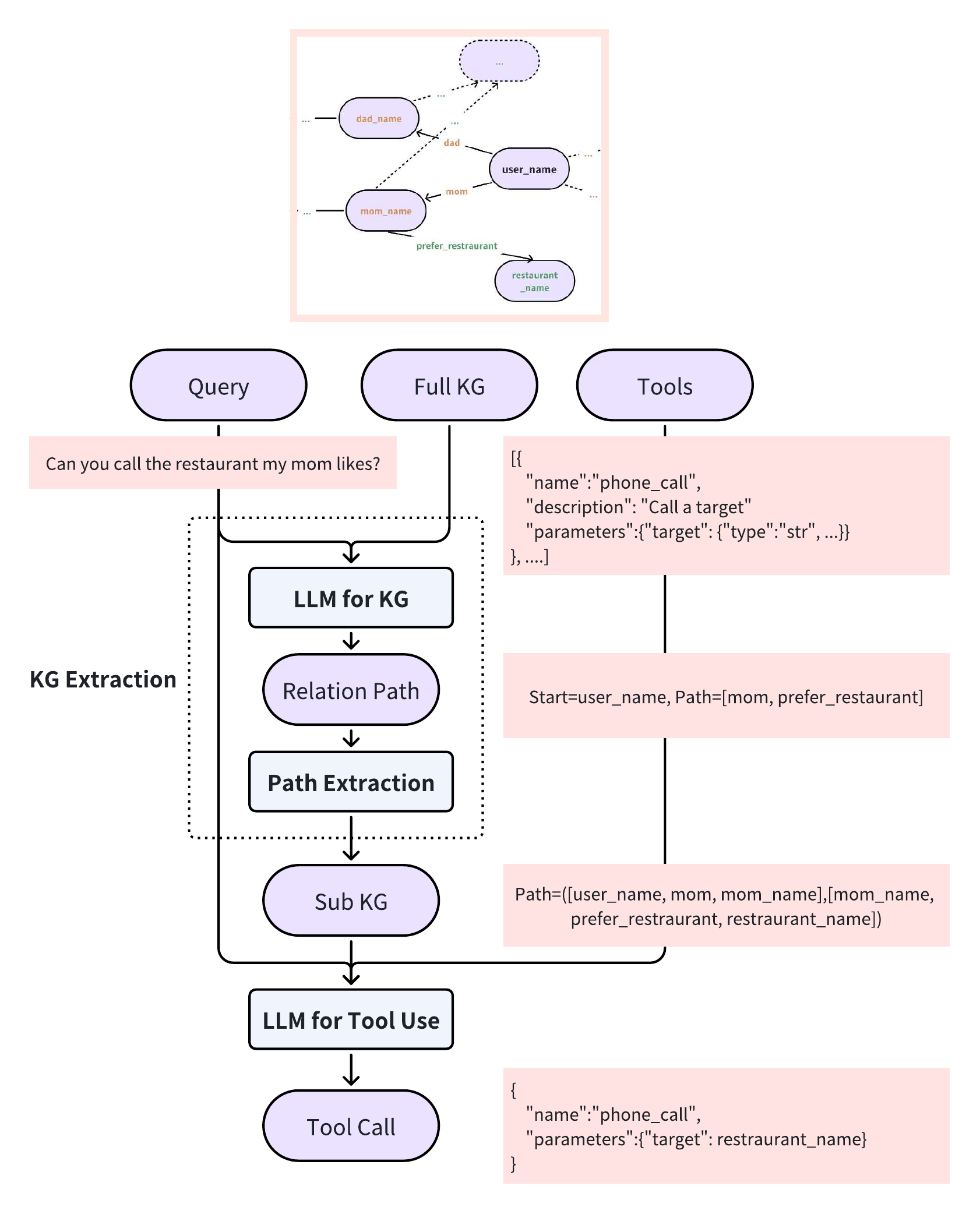}
    \caption{KGETool. We use two LLMs for each step respectively in KG extration and tool use.}
    \label{fig:simplepipeline}
\end{figure}

We propose KGETool for KG-augmented LLM Tool use to evaluate LLM's performances on FamilyTool, including 2 steps: KG Extraction and KG-augmented tool use, as shown in Figure \ref{fig:simplepipeline}. KG extraction generates a sub-KG relative to the query and LLM for tool use generates tool call based on the sub-KG. We design the pipeline for an extremely generalized scenario that the KG is totally inductive, since FamilyTool is designed for real-world personalized application, where KG is changing when the user is using his devices and new preferences are added with time. If the framework cannot handle the new KG without training or quickly tuning, it is meaningless for this framework to be evaluated in our benchmark. So for a simple start we choose to totally use the generalization ability of LLMs.

\subsection{KG Extraction}
First, we need to extract a sub-KG from the whole KG, which we call KG extraction. KG extraction is important because of two reasons. First, normally real-world KG is  so large that it cannot be put in the contexts of LLMs for understanding. Second, even though in our benchmark we design a relatively small KG that it can be put in LLM, the understanding ability of LLMs on KG is so limited~\cite{sui2024can} that the redundant information in KG can largely disturb LLMs. To further validate this, we made an ablation in Section \ref{ablation:fullkginllms} to show the bad performances of LLMs integrating with full KG. The KG Extraction mainly includes two steps: Relation Path Generation and Path Extraction.

\subsubsection{Relation Path Generation} 
We give the LLM the relations we have in the KG using prompt and use LLM to generate valid paths in KG to retrieve relevant information. In this way, we can easily handle the inductive setting that KG is changing with time because we can change the prompt we use for this path generation easily. Another reason we choose LLMs to generate paths is that we need the generalization ability of LLMs. The words in the query is miscellaneous compared to the relations we have in KG, so we want to use the LLM to generalize the word used in the query to the word we have in our KG. For example, the user use "child" in the query and our KG uses "son" in the relation, we want LLMs to generate the right relation "son" in our KG, not the fake relation "child". In path generation, we need LLMs to generate the output like "KG.search(Start=\textit{startentity}, Path=[\textit{relation1},\textit{relation2},...])" as shown in Figure \ref{fig:pathextraction} in Appendix. Then path extraction module will generate paths extracted from the KG.

\subsubsection{Path Extraction} 

When we got the relation path we need for our query, a simple way is to just use the start entity and relation path to get the target path using exact matching. A large problem of using LLMs to generating extraction paths is hallucination that LLMs can generate fake relations that are not in the KG. Using the same example before, we have "son" in our KG but sometimes in the query user could use "child". Even though we hope the generalization ability of LLM to resolve this problem and generate "son", unfortunately shown in our experiments in Section \ref{sec:kgextraction}, LLMs have a high probability of using "child" in path generation, which causes hallucination and error in extraction if using exact matching in KG. We can add "child" relation to resolve this problem at least, but this is just treating the symptoms but not the root cause. We cannot add all representation of one relation in our KG and we definitely will meet new relations in the inductive setting. So how to deal with fake relations LLM generates is an important question. We propose two alternative methods, Greedy Search (GS) and Relation Retrieval (RR) instead of common exact matching methods in path extraction. We show an example using these two methods in Figure \ref{fig:pathextraction} and demonstrate them below.

\textbf{Greedy Search.} In Greedy Search, we not only use exact matching for path extraction, but also when meeting fake relations, we will extract all links that starts with the previous end entity of last link. This can at least extract a larger sub-graph compared to aborting when meeting fake relations, and shrink the graph size compared to the whole KG graph. As in Figure \ref{fig:pathextraction}, "prefer\_alcohol\_store" is not a valid relation, so we will extract all links that start with "advisor\_name" which is the end entity of last link. In this way, assume we have $j$ fake relations in the path, we will have at most $\Pi_{i=1\text{ to }j} n_i$ paths where $n_i$ is the out degree of the fake relation start entity. Then we filter the wrong paths that don't exist in the KG.

\textbf{Relation Retrieval.} For all fake relations in the generated path, we use a well-trained retriever to get top\_$k$ valid relations in KG. $k$ here is a hyperparameter. And we use the Cartesian product to get new paths for extraction. Assume there are $j$ fake relations in one path, we can generate $k^j$ new paths conditioned on the original path and retriever. Then for those $k^j$ paths, we extract them respectively in the KG and drop wrong paths that don't exist in the KG.

\subsection{KG-augmented Tool Use}
With paths (sub-KG)  in hand, we can generally integrate them in LLMs for tool use. It was shown~\cite{sui2024can} that LLMs can understand paths best in the format of KG, so we give paths to LLM in the tool use step. For a query $Q$ and its accompanied paths $\mathcal{P}$, our input new query is "{$Q$}. The extra information for this query is ({$\mathcal{P}$})." And then the new query and candidate tools are all given to the LLM for tool use in a specific tool use format (if have) to generate tool call.

\section{Experiments}
We do experiments on FamilyTool-b and FamilyTool-e with KGETool. Results are shown separately.
 \subsection{Implementations}
\textbf{Baselines.} We choose seven different LLMs for evaluation, including five open-sourced LLMs from 7B to 32B: Qwen2.5-7B and 32B~\cite{qwen2}, Llama3.1-8B~\cite{dubey2024llama}, including the recently released Qwen3-8B and 32B~\cite{qwen3} and two large commercial LLMs: GPT-4o~\cite{OpenAI_hello_gpt4o} and DeepSeek-V3 (0324 version)~\cite{deepseekai2024deepseekv3technicalreport}. Besides, we also examine the performances of three LRMs (Large Reasoning Models) : QwQ-32B~\cite{qwq32b}, GPT-o3-mini~\cite{openai2025o3mini} and DeepSeek-r1~\cite{deepseekai2025deepseekr1incentivizingreasoningcapability}. We assume reasoning models can perform better in KG extraction which has an intrinsic logic of reasoning KG paths, so we use them in KG extraction and only QwQ-32B in tool use because o3-mini and deepseek-r1 do not have a tool-use template. We implement our experiments on A800 GPUs. For relation retrieval, we use jina-embeddings-v3~\cite{sturua2024jina} as the retriever. Further details are put in Appendix \ref{appendix:implementationdetails}.

\textbf{Metrics.} For KG extraction, we use EM, F1, No-Hallucination, Coverage and Format error to evaluate the results. No-Hallucination shows how many relation paths in KG extractionhave no fake relations. Coverage is to show whether the used KG entity of tool is covered in the sub-KG from KG extraction. Format error computes the ratio of queries that the model cannot generate valid KG search. For tool use, we use EM, Tool Acc., Value Acc.. EM shows if the functional calling is exactly true. Tool Acc. shows if the model can choose right tools from candidate tools. Value Acc. computes the accuracy of treating each parameter of a tool as a test sample separately. For results demonstration, we show the average value of the metrics across test set.

\subsection{KG Extraction}
\label{sec:kgextraction}
We show the results of KG extraction of different LLMs. We show the results of using Greedy Search in KG extraction because of better performances. Detailed comparisons of Greedy Search and Relation Retrieval are referred to Appendix \ref{appendix:ablation_gs_rr}.

\begin{table}[htbp]
    \centering

    \resizebox{0.48\textwidth}{!}{
        \begin{tabular}{l|ccccc|ccccc}
            \toprule
            & \multicolumn{5}{c}{FamilyTool-b}& \multicolumn{5}{c}{FamilyTool-e}
            \\
            \midrule
            LLM for KG & EM & F1 &No-Hal. & Coverage& Format Error & EM & F1 &No-Hal. & Coverage& Format Error
\\
            \midrule
            Llama3.1-8B & 43.89 & 59.33 & 52.42 &67.91 & 1.66 & 45.27 & 66.26 &60.73 & 59.56 & 3.74 
\\
            \midrule
            Qwen2.5-7B  & 43.89 & 60.72 & 51.47 &60.87 & 1.45 & 33.41 & 67.97 &62.03 & 42.42 & \textbf{0.44}\\
            
            Qwen2.5-32B  & 59.42 & 74.35 & 85.53 &68.53 &1.24 & 65.93 & 86.31 &94.04 & \textbf{71.43}& \textbf{0.44}\\

            QwQ-32B  & 59.21 & 74.24 & 87.10 &69.98 & 3.73 & \textbf{66.59}& 83.69 &92.87 & 71.21 & 4.40 
\\
            \midrule
             Qwen3-8B & 61.28 & 76.96 & 82.53 &75.57 & 1.66 & 61.76 & 83.80 &91.37 & 69.01 & 0.66 
\\

            Qwen3-32B & 58.39 & 75.90 &\textbf{95.39}&71.84 & 1.24 & 64.84 & 85.29 &91.59 & 70.99 & 0.66 
\\
            \midrule
            GPT-4o  & 61.90 & 76.82 &89.41 &73.08 & 2.28 & 62.42 & 83.95 &94.47 & 68.79 & 0.66 
\\

            DeepSeek-V3  & \textbf{65.42}&78.90 &92.68 &\textbf{75.78}& \textbf{1.04}& 62.64 & 84.08 &93.81 & 66.59 & 0.66 
\\
\midrule

            o3-mini  & 57.35 & 74.83 &93.90 &71.64 & 4.97 & 61.98 & 84.67 &93.38 & 68.79 & \textbf{0.44}\\

            DeepSeek-r1  & 65.01 &\textbf{79.38}&93.31 &74.53 & \textbf{1.04}& 65.05& \textbf{87.17}&\textbf{96.69}& 68.57 & \textbf{0.44}\\

            \bottomrule
        \end{tabular}
    }
    \caption{Results for LLM KG extraction using Greedy Search. No-Hal. denotes for No-Hallucination.}
    \label{tab:results_kgextraction}
\end{table}

We found that no LLMs can handle this task well, while reasoning models have no explicit advantages over non-reasoning models. In Qwen2 Series, 32B model can perform much better than 7B model while in Qwen3 Series, the advantage of 32B model only exists when there are more hops in FamilyTool-e. Also we found that the EM results on FamilyTool-e are similar to those on FamilyTool-b or even better, which shows that the increase of hop number in relationship does not increase the hardness of find the true path. However, the increase of hop number indeed increase the hardness of KG extraction, resulting in the decrease of coverage ratios in FamilyTool-e when retaining similar EM values for models like Qwen3-8B and commercial models. This is because LLM needs to generate more true hops to cover the true path in FamilyTool-e than FamilyTool-b because of more hop numbers.

\subsection{KG-augmented Tool use Generation}

\begin{table}[t]
    \centering
    \resizebox{0.48\textwidth}{!}{
        \begin{tabular}{l|ccc|ccc}
            \toprule
            & \multicolumn{3}{c}{FamilyTool-b}& \multicolumn{3}{c}{FamilyTool-e}
            \\
            \midrule
            LLM for Tool Use & EM &Tool Acc. & Value Acc. & EM &Tool Acc. & Value Acc.\\
            \midrule
            Llama3.1-8B & 30.85 & 67.08 & 38.90  & 30.77 &51.43 & 34.78 
\\
            \midrule
            Qwen2.5-7B  & 47.00 & 81.37 & 55.63  & 52.97 &81.10 & 61.62 
\\
            Qwen2.5-32B  & 42.03 &64.18 & 48.37  & 39.56 &63.07 & 46.40 
\\
            QwQ-32B  & 50.72 &  \textbf{82.40}&59.55  & 52.97&78.90 & 61.16 
\\
            \midrule
            Qwen3-8B  & \textbf{53.00}& 80.75 &\textbf{61.38}& 55.82 &80.65 & 63.94 
\\

            Qwen3-32B& 51.55 & 81.37 &60.18  & \textbf{56.92}&81.54 & \textbf{65.10}\\
                        \midrule
            GPT-4o  & 26.29 & 75.36 &29.99  & 36.26 &\textbf{81.75}& 40.44 
\\
             DeepSeek-V3  & 42.03 & 67.08 &50.09  & 44.39 &63.29 & 51.91 \\
            \bottomrule
        \end{tabular}
          
    }
      \caption{Results for LLM tool use with Golden sub-KG.}
    \label{tab:results_goldenkg_tooluse}
\end{table}
\textbf{Golden KG.} We first show the KG-augmented Tool-use Generation results of golden KG. We found that even with golden sub-KGs, the LLMs cannot achieve good performances while the best model achieves only above 50. And reasoning models can get better performances compared to non-reasoning ones in tool use when comparing Qwen2.5-32B and QwQ. The low performances of of Qwen2.5-32B (even lower than 7B) is due to the low ratio of tool call, which is 75.98 compared to 97.10 in Qwen2.5-7B. After checking the bad cases, we found that Qwen2.5-32B tends to ask for a multi-round conversation instead of applying tool use, because of hallucination in the parameters that are required for tool use, while 7B model will enforce tool use even with hallucination. GPT-4o performs badly because it cannot identify the entity in our KG as a valid value for the parameter. Similarly to KG extraction, the results of FamilyTool-e are similar to that of FamilyTool-b for the same LLM, showing that the more hop numbers in the true path will not make it harder for LLMs, which is easy to explain in that LLMs only need to use the last entity of the path no matter how long is the path.

\textbf{Extracted KG.} Then we show the results using the extracted KG in Table \ref{tab:results_pipeline_tooluse}. Like the results with golden KG, Qwen3 Series models get the best performances but as assumed, the low coverage ratio of KG extraction will definitely downgrade the tool use performances and the EM values are near 40 even for the best LLMs. 

\begin{table}[htbp]
    \centering

    \resizebox{0.48\textwidth}{!}{
        \begin{tabular}{l|ccc|ccc}
            \toprule
                        & \multicolumn{3}{c}{FamilyTool-b}& \multicolumn{3}{c}{FamilyTool-e}
            \\
            \midrule
            LLM for Tool Use& EM &Tool Acc. & Value Acc.& EM &Tool Acc. & Value Acc.\\
            \midrule
            Llama3.1-8B & 16.36 & 54.66 & 23.89 & 15.60 &44.62 & 20.42 
\\     \midrule
            Qwen2.5-7B  & 28.78 & 78.88 & 38.19 & 16.70 &75.38 & 26.67 
\\
             Qwen2.5-32B  & 28.99 &67.08 & 38.28 & 33.63 &70.11 & 42.41 
\\
            QwQ-32B  & 37.06 & 79.30 &45.47 & 39.56 &75.38 & 47.94 
\\     \midrule
            Qwen3-8B  & \textbf{40.37} & \textbf{80.33} &\textbf{49.23} & 38.02 &78.90 & 47.89 
\\
       
            Qwen3-32B& 36.02 & \textbf{80.33}&46.06 & \textbf{39.78} &\textbf{79.12} & \textbf{48.89} 
\\     \midrule
            GPT-4o  & 19.67 & 76.81 &23.64 & 23.74 &78.68 & 28.30 
\\
             DeepSeek-V3  & 30.43 & 64.60 &39.23 & 27.91 &58.90 & 36.53 \\
            \bottomrule
        \end{tabular}
           
    }
 \caption{Results for LLM tool use with Extracted sub-KG with Greedy Search.}
    \label{tab:results_pipeline_tooluse}
\end{table}

\section{Ablations}
In ablations in the main paper, we show the importance of using Knowledge Graphs in Multi-hop scenerios compared with using documents with normal document retriever, which is important in KG-based tasks. Due to page limitation, the left ablations, including comparison of Greedy Search and Relation Retrieval in KG extraction, usage of full KG and real KG values in Tool use, and combining different LLMs in two steps can get better performances are put in Appendix \ref{appendix:ablations_more}.
\subsection{KG vs Documents}
\begin{figure}
    \centering
    \includegraphics[width=\linewidth]{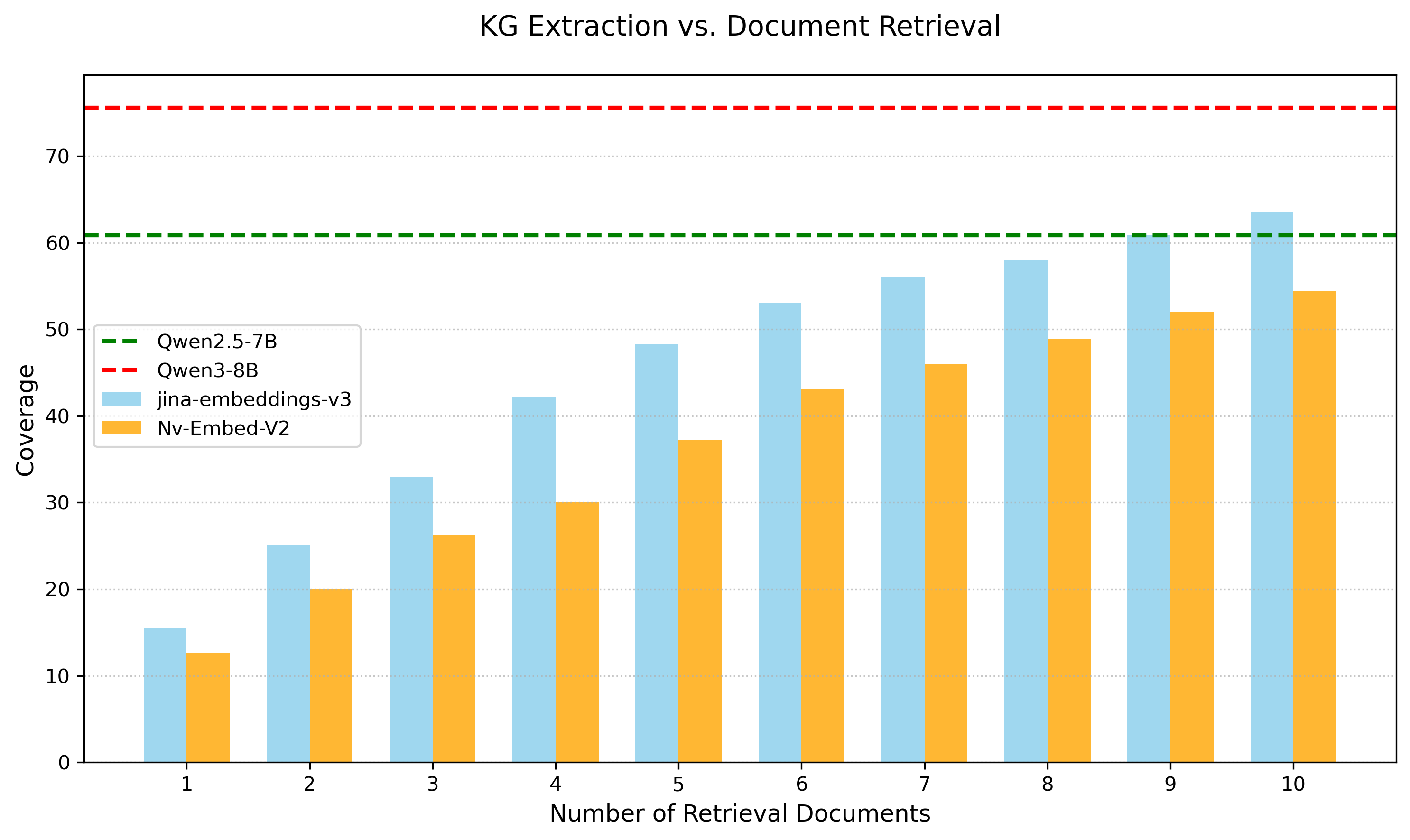}
    \caption{KG extraction vs. Documents Retrieval. We show the coverage ratio of document retrievers with 1 to 10 retrieved documents.}
    \label{fig:ablation_KGvsDocuments_extraction}
\end{figure}

\begin{figure}
    \centering
    \includegraphics[width=\linewidth]{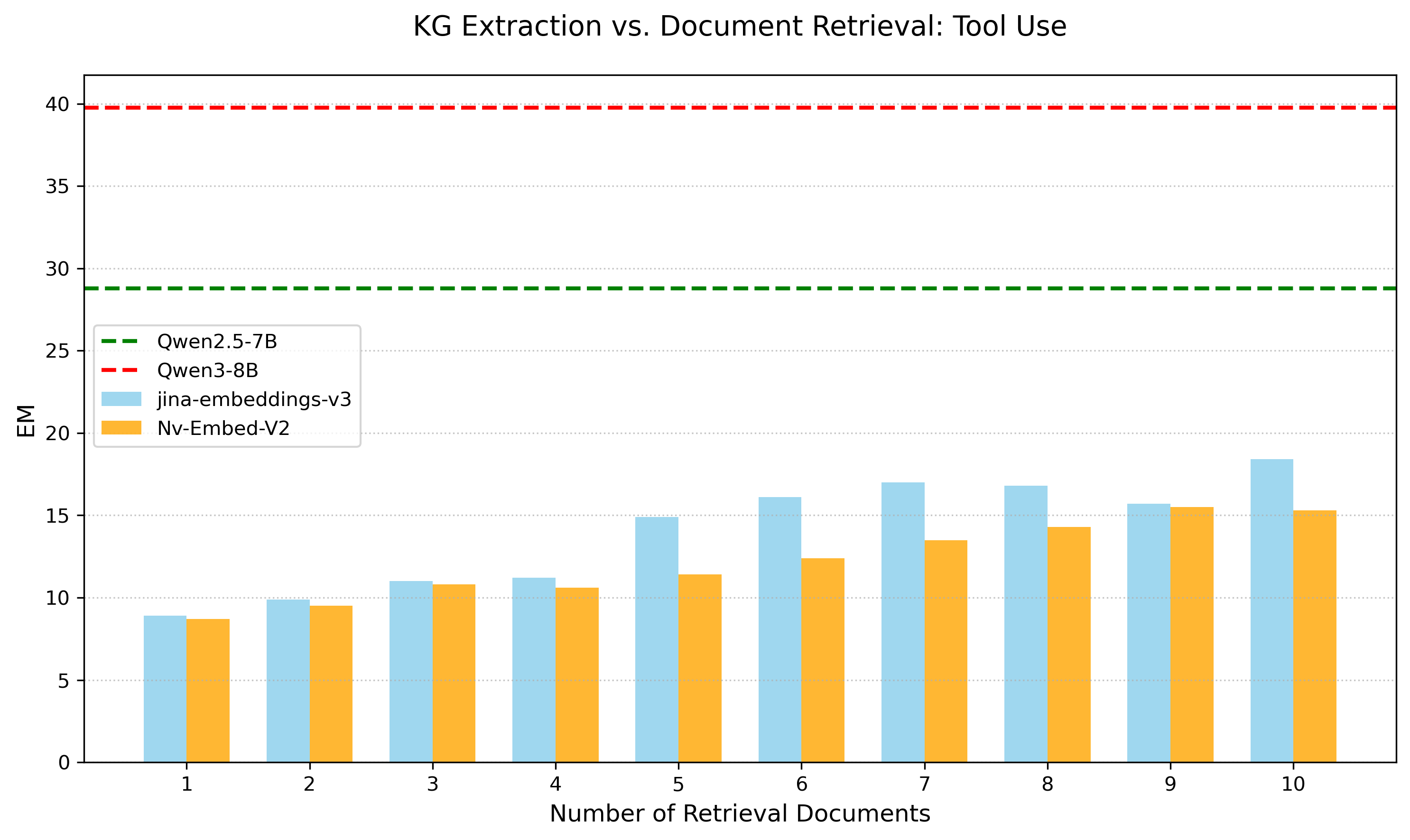}
    \caption{KG extraction vs. Documents Retreival on Tool use. We use Qwen2.5-7B for the tool use model.}
    \label{fig:ablation_KGvsDocuments_tooluse}
    \vspace{-3mm}
\end{figure}

We transform all KG links into documents and then use retriever for the input query to get documents. To evaluate the retrieval performances, we re-transform the retrieved documents into links and evaluate them like that in Table \ref{tab:results_kgextraction}. We choose jina-embedding-V3 and Nv-Embed-V2~\cite{moreira2024nv,lee2024nv} for retrieval for documents with up to 6 documents. Results are shown and compared with Qwen2.5-7B and Qwen3-8B in KG extraction in Figure \ref{fig:ablation_KGvsDocuments_extraction}. Also, we evaluate differenct Retrieval methods' performances in Tool use with Qwen2.5-7B model in Figure \ref{fig:ablation_KGvsDocuments_tooluse}. We found that although jina can achieve similar performances of coverage in document extraction with 7B model when retrieving up to ten documents, using those documents in tool use failed to achieve comparable results with KG extraction. 

Two plausible explanations may account for this. First retrieving ten documents will introduce more noises in the candidates that will make LLMs harder for tool use, which can be shown in Figure \ref{fig:ablation_KGvsDocuments_tooluse} that jina model performance decreases when retrieval documents increase from 7 to 8, 9. The same phenomenon appears in Nv model when retrieval documents increase from 3 to 4 and 9 to 10. Instead, KG extraction has a high ratio of EM samples when obtaining the similar coverage ratio with document retrieval, that will only give LLM knowledge that are needed without more noise. Also, document retrieval has no logic helping the reasoning of LLMs while KG extraction gives LLMs a reasoning path that helps much. In the end, common retriever cannot solve multi-hop relational retrieval in our scenerio compared to KG extraction, which shows the importance of using Knowledge Graphs for retrieval.


\section{Conclusion}
In this paper, we introduce a new multi-hop personalized tool use scenario for family-specific tool use in the inductive setting and propose a new benchmark: FamilyTool, including base dataset FamilyTool-b and extended one FamilyTool-e, differentiated in how many relationship hops are needed for query. With GPT-4o generation and human check, our benchmark includes multi-hop queries with different difficulties and hops. To test LLMs' tool use ability , we propose KGETool, a simple pipeline for the inductive KG setting. And all evaluations in the pipeline are based on the ground truth with ruled-based metrics without using any GPT or human evaluation. Experiments show that existing LLMs are far less strong enough for this real-world driven application, while recently released Qwen3 Series perform the best. So we hope FamilyTool can guide researchers in this new, unearthed domain in tool use of LLMs. 

\section{Limitations}
Our experiments are constrained to a predefined set of tools and scenarios with totally near one thousand samples, which may not fully capture the diversity of real-world tool-use tasks. Additionally, our evaluation focuses primarily on static tool interactions, whereas dynamic, multi-turn tool use in open-ended environments presents further challenges. 

\section{Ethic Statement}

\textbf{Potential Risks.}
Regarding broader societal impact, there is of course some risk that better personalized tool use could be used for nefarious purposes, such as recommending harmful content to minors.

All APIs used in our benchmark were cited from previous works and good citation is added. The datas included in our benchmark were created and checked by the authors only. Full text of instructions, recruitment and payment, data consent are given to participants. The data collection protocal was approved by an ethics review board. AI assistants are used for helping to decorate the paper.


\bibliography{custom,tooluse,inductivekg,multihopqa}

\begin{thebibliography}{82}
\providecommand{\natexlab}[1]{#1}

\bibitem[{Basu et~al.(2024)Basu, Abdelaziz, Chaudhury, Dan, Crouse, Munawar, Austel, Kumaravel, Muthusamy, Kapanipathi, and Lastras}]{basu_api-blend_2024_1}
Kinjal Basu, Ibrahim Abdelaziz, Subhajit Chaudhury, Soham Dan, Maxwell Crouse, Asim Munawar, Vernon Austel, Sadhana Kumaravel, Vinod Muthusamy, Pavan Kapanipathi, and Luis Lastras. 2024.
\newblock \href {https://doi.org/10.18653/v1/2024.acl-long.694} {{API}-{BLEND}: A comprehensive corpora for training and benchmarking {API} {LLM}s}.
\newblock In \emph{Proceedings of the 62nd Annual Meeting of the Association for Computational Linguistics (Volume 1: Long Papers)}, pages 12859--12870, Bangkok, Thailand. Association for Computational Linguistics.

\bibitem[{Berant et~al.(2013)Berant, Chou, Frostig, and Liang}]{berant2013semantic}
Jonathan Berant, Andrew Chou, Roy Frostig, and Percy Liang. 2013.
\newblock Semantic parsing on freebase from question-answer pairs.
\newblock In \emph{Proceedings of the 2013 conference on empirical methods in natural language processing}, pages 1533--1544.

\bibitem[{Bordes et~al.(2013)Bordes, Usunier, Garc{\'{\i}}a{-}Dur{\'{a}}n, Weston, and Yakhnenko}]{DBLP:conf/nips/BordesUGWY13}
Antoine Bordes, Nicolas Usunier, Alberto Garc{\'{\i}}a{-}Dur{\'{a}}n, Jason Weston, and Oksana Yakhnenko. 2013.
\newblock \href {https://proceedings.neurips.cc/paper/2013/hash/1cecc7a77928ca8133fa24680a88d2f9-Abstract.html} {Translating embeddings for modeling multi-relational data}.
\newblock In \emph{Advances in Neural Information Processing Systems 26: 27th Annual Conference on Neural Information Processing Systems 2013. Proceedings of a meeting held December 5-8, 2013, Lake Tahoe, Nevada, United States}, pages 2787--2795.

\bibitem[{Chen et~al.(2024)Chen, Du, Zhang, Liu, Liu, Zheng, Zhuo, Zhang, Lin, Chen, and Zhao}]{chen_t-eval_2024_1}
Zehui Chen, Weihua Du, Wenwei Zhang, Kuikun Liu, Jiangning Liu, Miao Zheng, Jingming Zhuo, Songyang Zhang, Dahua Lin, Kai Chen, and Feng Zhao. 2024.
\newblock \href {https://doi.org/10.18653/v1/2024.acl-long.515} {{T}-eval: Evaluating the tool utilization capability of large language models step by step}.
\newblock In \emph{Proceedings of the 62nd Annual Meeting of the Association for Computational Linguistics (Volume 1: Long Papers)}, pages 9510--9529, Bangkok, Thailand. Association for Computational Linguistics.

\bibitem[{Cui et~al.(2023)Cui, Wang, Sun, Liu, Jiang, Han, and Hu}]{cui2023lifelong}
Yuanning Cui, Yuxin Wang, Zequn Sun, Wenqiang Liu, Yiqiao Jiang, Kexin Han, and Wei Hu. 2023.
\newblock Lifelong embedding learning and transfer for growing knowledge graphs.
\newblock In \emph{Proceedings of the AAAI Conference on Artificial Intelligence}, volume~37, pages 4217--4224.

\bibitem[{Das et~al.(2018)Das, Dhuliawala, Zaheer, Vilnis, Durugkar, Krishnamurthy, Smola, and McCallum}]{DBLP:conf/iclr/DasDZVDKSM18}
Rajarshi Das, Shehzaad Dhuliawala, Manzil Zaheer, Luke Vilnis, Ishan Durugkar, Akshay Krishnamurthy, Alex Smola, and Andrew McCallum. 2018.
\newblock Go for a walk and arrive at the answer: Reasoning over paths in knowledge bases using reinforcement learning.
\newblock In \emph{{ICLR} (Poster)}. OpenReview.net.

\bibitem[{Das et~al.(2017)Das, Neelakantan, Belanger, and McCallum}]{DBLP:conf/eacl/McCallumNDB17}
Rajarshi Das, Arvind Neelakantan, David Belanger, and Andrew McCallum. 2017.
\newblock Chains of reasoning over entities, relations, and text using recurrent neural networks.
\newblock In \emph{{EACL} {(1)}}, pages 132--141. Association for Computational Linguistics.

\bibitem[{DeepSeek-AI(2024)}]{deepseekai2024deepseekv3technicalreport}
DeepSeek-AI. 2024.
\newblock \href {https://arxiv.org/abs/2412.19437} {Deepseek-v3 technical report}.
\newblock \emph{Preprint}, arXiv:2412.19437.

\bibitem[{DeepSeek-AI(2025)}]{deepseekai2025deepseekr1incentivizingreasoningcapability}
DeepSeek-AI. 2025.
\newblock \href {https://arxiv.org/abs/2501.12948} {Deepseek-r1: Incentivizing reasoning capability in llms via reinforcement learning}.
\newblock \emph{Preprint}, arXiv:2501.12948.

\bibitem[{Dettmers et~al.(2018)Dettmers, Pasquale, Pontus, and Riedel}]{dettmers2018conve}
Tim Dettmers, Minervini Pasquale, Stenetorp Pontus, and Sebastian Riedel. 2018.
\newblock Convolutional 2d knowledge graph embeddings.
\newblock In \emph{AAAI}.

\bibitem[{Dubey et~al.(2024{\natexlab{a}})Dubey, Jauhri, Pandey, Kadian, Al-Dahle, Letman, Mathur, Schelten, Yang, Fan et~al.}]{dubey2024llama}
Abhimanyu Dubey, Abhinav Jauhri, Abhinav Pandey, Abhishek Kadian, Ahmad Al-Dahle, Aiesha Letman, Akhil Mathur, Alan Schelten, Amy Yang, Angela Fan, and 1 others. 2024{\natexlab{a}}.
\newblock The llama 3 herd of models.
\newblock \emph{arXiv preprint arXiv:2407.21783}.

\bibitem[{Dubey et~al.(2024{\natexlab{b}})Dubey, Jauhri, Pandey, Kadian, Al-Dahle, Letman, Mathur, Zhao, and et~al.}]{dubey_llama_2024_1}
Abhimanyu Dubey, Abhinav Jauhri, Abhinav Pandey, Abhishek Kadian, Ahmad Al-Dahle, Aiesha Letman, Akhil Mathur, Zhiwei Zhao, and et~al. 2024{\natexlab{b}}.
\newblock \href {http://arxiv.org/abs/2407.21783} {The {Llama} 3 {Herd} of {Models}}.
\newblock \emph{arXiv preprint}.
\newblock ArXiv:2407.21783 [cs].

\bibitem[{Farn and Shin(2023)}]{farn_tooltalk_2023}
Nicholas Farn and Richard Shin. 2023.
\newblock \href {https://doi.org/10.48550/arXiv.2311.10775} {{ToolTalk}: {Evaluating} {Tool}-{Usage} in a {Conversational} {Setting}}.
\newblock \emph{arXiv preprint}.
\newblock ArXiv:2311.10775 [cs].

\bibitem[{Gal{\'{a}}rraga et~al.(2015)Gal{\'{a}}rraga, Teflioudi, Hose, and Suchanek}]{DBLP:journals/vldb/GalarragaTHS15}
Luis Gal{\'{a}}rraga, Christina Teflioudi, Katja Hose, and Fabian~M. Suchanek. 2015.
\newblock Fast rule mining in ontological knowledge bases with {AMIE+}.
\newblock \emph{{VLDB} J.}, 24(6):707--730.

\bibitem[{Galkin et~al.(2022)Galkin, Denis, Wu, and Hamilton}]{DBLP:conf/iclr/0001DWH22}
Mikhail Galkin, Etienne~G. Denis, Jiapeng Wu, and William~L. Hamilton. 2022.
\newblock \href {https://openreview.net/forum?id=xMJWUKJnFSw} {Nodepiece: Compositional and parameter-efficient representations of large knowledge graphs}.
\newblock In \emph{The Tenth International Conference on Learning Representations, {ICLR} 2022, Virtual Event, April 25-29, 2022}. OpenReview.net.

\bibitem[{Gao et~al.(2023)Gao, Madaan, Zhou, Alon, Liu, Yang, Callan, and Neubig}]{gao_pal_2022_1}
Luyu Gao, Aman Madaan, Shuyan Zhou, Uri Alon, Pengfei Liu, Yiming Yang, Jamie Callan, and Graham Neubig. 2023.
\newblock \href {https://proceedings.mlr.press/v202/gao23f.html} {{PAL}: Program-aided language models}.
\newblock In \emph{Proceedings of the 40th International Conference on Machine Learning}, volume 202 of \emph{Proceedings of Machine Learning Research}, pages 10764--10799. PMLR.

\bibitem[{Geng et~al.(2023)Geng, Chen, Pan, Chen, Jiang, Zhang, and Chen}]{geng2023relational}
Yuxia Geng, Jiaoyan Chen, Jeff~Z Pan, Mingyang Chen, Song Jiang, Wen Zhang, and Huajun Chen. 2023.
\newblock Relational message passing for fully inductive knowledge graph completion.
\newblock In \emph{2023 IEEE 39th international conference on data engineering (ICDE)}, pages 1221--1233. IEEE.

\bibitem[{Guo et~al.(2024)Guo, Cheng, Wang, Liang, Qin, Li, Liu, Sun, and Liu}]{guo_stabletoolbench_2024}
Zhicheng Guo, Sijie Cheng, Hao Wang, Shihao Liang, Yujia Qin, Peng Li, Zhiyuan Liu, Maosong Sun, and Yang Liu. 2024.
\newblock \href {http://arxiv.org/abs/2403.07714} {{StableToolBench}: {Towards} {Stable} {Large}-{Scale} {Benchmarking} on {Tool} {Learning} of {Large} {Language} {Models}}.
\newblock \emph{arXiv preprint}.
\newblock ArXiv:2403.07714 [cs].

\bibitem[{Hao et~al.(2025)Hao, Cao, Jin, Liao, Chen, Liu, and Zhao}]{hao2025evaluatingpersonalizedtoolaugmentedllms}
Yupu Hao, Pengfei Cao, Zhuoran Jin, Huanxuan Liao, Yubo Chen, Kang Liu, and Jun Zhao. 2025.
\newblock \href {https://arxiv.org/abs/2503.00771} {Evaluating personalized tool-augmented llms from the perspectives of personalization and proactivity}.
\newblock \emph{Preprint}, arXiv:2503.00771.

\bibitem[{Ho et~al.(2020)Ho, Duong~Nguyen, Sugawara, and Aizawa}]{ho-etal-2020-constructing}
Xanh Ho, Anh-Khoa Duong~Nguyen, Saku Sugawara, and Akiko Aizawa. 2020.
\newblock \href {https://doi.org/10.18653/v1/2020.coling-main.580} {Constructing a multi-hop {QA} dataset for comprehensive evaluation of reasoning steps}.
\newblock In \emph{Proceedings of the 28th International Conference on Computational Linguistics}, pages 6609--6625, Barcelona, Spain (Online). International Committee on Computational Linguistics.

\bibitem[{Huang et~al.(2024{\natexlab{a}})Huang, Zhong, Lu, Zhu, Gao, Liu, Hou, Zeng, Wang, Shang, Jiang, Xu, and Liu}]{huang_planning_2024_1}
Shijue Huang, Wanjun Zhong, Jianqiao Lu, Qi~Zhu, Jiahui Gao, Weiwen Liu, Yutai Hou, Xingshan Zeng, Yasheng Wang, Lifeng Shang, Xin Jiang, Ruifeng Xu, and Qun Liu. 2024{\natexlab{a}}.
\newblock \href {https://doi.org/10.18653/v1/2024.findings-acl.259} {Planning, creation, usage: Benchmarking {LLM}s for comprehensive tool utilization in real-world complex scenarios}.
\newblock In \emph{Findings of the Association for Computational Linguistics: ACL 2024}, pages 4363--4400, Bangkok, Thailand. Association for Computational Linguistics.

\bibitem[{Huang et~al.(2024{\natexlab{b}})Huang, Shi, Li, Fan, Wu, Zhang, Liu, Zhou, Wan, Gong, and Sun}]{huang_metatool_2024_1}
Yue Huang, Jiawen Shi, Yuan Li, Chenrui Fan, Siyuan Wu, Qihui Zhang, Yixin Liu, Pan Zhou, Yao Wan, Neil~Zhenqiang Gong, and Lichao Sun. 2024{\natexlab{b}}.
\newblock \href {https://openreview.net/forum?id=R0c2qtalgG} {Metatool benchmark for large language models: Deciding whether to use tools and which to use}.
\newblock In \emph{The Twelfth International Conference on Learning Representations}.

\bibitem[{Kwon et~al.(2023)Kwon, Li, Zhuang, Sheng, Zheng, Yu, Gonzalez, Zhang, and Stoica}]{kwon2023efficient}
Woosuk Kwon, Zhuohan Li, Siyuan Zhuang, Ying Sheng, Lianmin Zheng, Cody~Hao Yu, Joseph~E. Gonzalez, Hao Zhang, and Ion Stoica. 2023.
\newblock Efficient memory management for large language model serving with pagedattention.
\newblock In \emph{Proceedings of the ACM SIGOPS 29th Symposium on Operating Systems Principles}.

\bibitem[{Lee et~al.(2024)Lee, Roy, Xu, Raiman, Shoeybi, Catanzaro, and Ping}]{lee2024nv}
Chankyu Lee, Rajarshi Roy, Mengyao Xu, Jonathan Raiman, Mohammad Shoeybi, Bryan Catanzaro, and Wei Ping. 2024.
\newblock Nv-embed: Improved techniques for training llms as generalist embedding models.
\newblock \emph{arXiv preprint arXiv:2405.17428}.

\bibitem[{Lee et~al.(2023)Lee, Chung, and Whang}]{pmlr-v202-lee23c}
Jaejun Lee, Chanyoung Chung, and Joyce~Jiyoung Whang. 2023.
\newblock \href {https://proceedings.mlr.press/v202/lee23c.html} {{I}n{G}ram: Inductive knowledge graph embedding via relation graphs}.
\newblock In \emph{Proceedings of the 40th International Conference on Machine Learning}, volume 202 of \emph{Proceedings of Machine Learning Research}, pages 18796--18809. PMLR.

\bibitem[{Li et~al.(2023)Li, Zhao, Yu, Song, Li, Yu, Li, Huang, and Li}]{li_api-bank_2023_1}
Minghao Li, Yingxiu Zhao, Bowen Yu, Feifan Song, Hangyu Li, Haiyang Yu, Zhoujun Li, Fei Huang, and Yongbin Li. 2023.
\newblock \href {https://doi.org/10.18653/v1/2023.emnlp-main.187} {{API}-bank: A comprehensive benchmark for tool-augmented {LLM}s}.
\newblock In \emph{Proceedings of the 2023 Conference on Empirical Methods in Natural Language Processing}, pages 3102--3116, Singapore. Association for Computational Linguistics.

\bibitem[{Liu et~al.(2021)Liu, Grau, Horrocks, and Kostylev}]{DBLP:conf/nips/LiuGHK21}
Shuwen Liu, Bernardo~Cuenca Grau, Ian Horrocks, and Egor~V. Kostylev. 2021.
\newblock {INDIGO:} gnn-based inductive knowledge graph completion using pair-wise encoding.
\newblock In \emph{NeurIPS}, pages 2034--2045.

\bibitem[{Liu et~al.(2024)Liu, Huang, Zeng, Hao, Yu, Li, Wang, Gan, Liu, Yu, Wang, Wang, Ning, Hou, Wang, Wu, Wang, Liu, Wang, Tang, Tu, Shang, Jiang, Tang, Lian, Liu, and Chen}]{liu_toolace_2024}
Weiwen Liu, Xu~Huang, Xingshan Zeng, Xinlong Hao, Shuai Yu, Dexun Li, Shuai Wang, Weinan Gan, Zhengying Liu, Yuanqing Yu, Zezhong Wang, Yuxian Wang, Wu~Ning, Yutai Hou, Bin Wang, Chuhan Wu, Xinzhi Wang, Yong Liu, Yasheng Wang, and 8 others. 2024.
\newblock \href {http://arxiv.org/abs/2409.00920} {{ToolACE}: {Winning} the {Points} of {LLM} {Function} {Calling}}.
\newblock \emph{arXiv preprint}.
\newblock ArXiv:2409.00920 [cs].

\bibitem[{Lu et~al.(2024)Lu, Holleis, Zhang, Aumayer, Nan, Bai, Ma, Ma, Li, Yin, Wang, and Pang}]{lu_toolsandbox_2024}
Jiarui Lu, Thomas Holleis, Yizhe Zhang, Bernhard Aumayer, Feng Nan, Felix Bai, Shuang Ma, Shen Ma, Mengyu Li, Guoli Yin, Zirui Wang, and Ruoming Pang. 2024.
\newblock \href {http://arxiv.org/abs/2408.04682} {{ToolSandbox}: {A} {Stateful}, {Conversational}, {Interactive} {Evaluation} {Benchmark} for {LLM} {Tool} {Use} {Capabilities}}.
\newblock \emph{arXiv preprint}.
\newblock ArXiv:2408.04682 [cs].

\bibitem[{Ma et~al.(2024)Ma, Gou, Hao, Xu, Wang, Pan, Yang, Cao, Sun, Awadalla, and Chen}]{ma_sciagent_2024}
Yubo Ma, Zhibin Gou, Junheng Hao, Ruochen Xu, Shuohang Wang, Liangming Pan, Yujiu Yang, Yixin Cao, Aixin Sun, Hany Awadalla, and Weizhu Chen. 2024.
\newblock \href {http://arxiv.org/abs/2402.11451} {{SciAgent}: {Tool}-augmented {Language} {Models} for {Scientific} {Reasoning}}.
\newblock \emph{arXiv preprint}.
\newblock ArXiv:2402.11451 [cs].

\bibitem[{Mai et~al.(2021)Mai, Zheng, Yang, and Hu}]{DBLP:conf/aaai/MaiZY021}
Sijie Mai, Shuangjia Zheng, Yuedong Yang, and Haifeng Hu. 2021.
\newblock Communicative message passing for inductive relation reasoning.
\newblock In \emph{{AAAI}}, pages 4294--4302. {AAAI} Press.

\bibitem[{Mallen et~al.(2022)Mallen, Asai, Zhong, Das, Khashabi, and Hajishirzi}]{mallen2022not}
Alex Mallen, Akari Asai, Victor Zhong, Rajarshi Das, Daniel Khashabi, and Hannaneh Hajishirzi. 2022.
\newblock When not to trust language models: Investigating effectiveness of parametric and non-parametric memories.
\newblock \emph{arXiv preprint arXiv:2212.10511}.

\bibitem[{Meilicke et~al.(2019)Meilicke, Chekol, Ruffinelli, and Stuckenschmidt}]{DBLP:conf/ijcai/MeilickeCRS19}
Christian Meilicke, Melisachew~Wudage Chekol, Daniel Ruffinelli, and Heiner Stuckenschmidt. 2019.
\newblock \href {https://doi.org/10.24963/ijcai.2019/435} {Anytime bottom-up rule learning for knowledge graph completion}.
\newblock In \emph{Proceedings of the Twenty-Eighth International Joint Conference on Artificial Intelligence, {IJCAI} 2019, Macao, China, August 10-16, 2019}, pages 3137--3143. ijcai.org.

\bibitem[{Moreira et~al.(2024)Moreira, Osmulski, Xu, Ak, Schifferer, and Oldridge}]{moreira2024nv}
Gabriel de Souza~P Moreira, Radek Osmulski, Mengyao Xu, Ronay Ak, Benedikt Schifferer, and Even Oldridge. 2024.
\newblock Nv-retriever: Improving text embedding models with effective hard-negative mining.
\newblock \emph{arXiv preprint arXiv:2407.15831}.

\bibitem[{Neelakantan et~al.(2015)Neelakantan, Roth, and McCallum}]{DBLP:conf/acl/NeelakantanRM15}
Arvind Neelakantan, Benjamin Roth, and Andrew McCallum. 2015.
\newblock Compositional vector space models for knowledge base completion.
\newblock In \emph{{ACL} {(1)}}, pages 156--166. The Association for Computer Linguistics.

\bibitem[{Omran et~al.(2018)Omran, Wang, and Wang}]{DBLP:conf/ijcai/OmranWW18}
Pouya~Ghiasnezhad Omran, Kewen Wang, and Zhe Wang. 2018.
\newblock Scalable rule learning via learning representation.
\newblock In \emph{{IJCAI}}, pages 2149--2155. ijcai.org.

\bibitem[{OpenAI()}]{OpenAI_hello_gpt4o}
OpenAI.
\newblock Hello gpt-4o.
\newblock \url{https://openai.com/index/hello-gpt-4o/}.
\newblock [Online; accessed 22-January-2025].

\bibitem[{{OpenAI}(2025)}]{openai2025o3mini}
{OpenAI}. 2025.
\newblock \href {https://openai.com/index/openai-o3-mini/} {Openai o3-mini}.
\newblock Released on January 31, 2025.

\bibitem[{Pan et~al.(2024)Pan, Cao, Lin, Han, Zheng, Wang, Cai, and Sun}]{pan2024not}
Ruotong Pan, Boxi Cao, Hongyu Lin, Xianpei Han, Jia Zheng, Sirui Wang, Xunliang Cai, and Le~Sun. 2024.
\newblock Not all contexts are equal: Teaching llms credibility-aware generation.
\newblock \emph{arXiv preprint arXiv:2404.06809}.

\bibitem[{Patil et~al.(2023)Patil, Zhang, Wang, and Gonzalez}]{patil_gorilla_2023}
Shishir~G. Patil, Tianjun Zhang, Xin Wang, and Joseph~E. Gonzalez. 2023.
\newblock \href {https://doi.org/10.48550/arXiv.2305.15334} {Gorilla: {Large} {Language} {Model} {Connected} with {Massive} {APIs}}.
\newblock \emph{arXiv preprint}.
\newblock ArXiv:2305.15334 [cs].

\bibitem[{Pirr{\`{o}}(2020)}]{DBLP:conf/aaai/Pirro20}
Giuseppe Pirr{\`{o}}. 2020.
\newblock Relatedness and tbox-driven rule learning in large knowledge bases.
\newblock In \emph{{AAAI}}, pages 2975--2982. {AAAI} Press.

\bibitem[{Qin et~al.(2023)Qin, Liang, Ye, Zhu, Yan, Lu, Lin, Cong, Tang, Qian, Zhao, Hong, Tian, Xie, Zhou, Gerstein, Li, Liu, and Sun}]{qin_toolllm_2023}
Yujia Qin, Shihao Liang, Yining Ye, Kunlun Zhu, Lan Yan, Yaxi Lu, Yankai Lin, Xin Cong, Xiangru Tang, Bill Qian, Sihan Zhao, Lauren Hong, Runchu Tian, Ruobing Xie, Jie Zhou, Mark Gerstein, Dahai Li, Zhiyuan Liu, and Maosong Sun. 2023.
\newblock \href {http://arxiv.org/abs/2307.16789} {{ToolLLM}: {Facilitating} {Large} {Language} {Models} to {Master} 16000+ {Real}-world {APIs}}.
\newblock \emph{arXiv preprint}.
\newblock ArXiv:2307.16789 [cs].

\bibitem[{Qu et~al.(2021)Qu, Chen, Xhonneux, Bengio, and Tang}]{DBLP:conf/iclr/QuCXBT21}
Meng Qu, Junkun Chen, Louis{-}Pascal A.~C. Xhonneux, Yoshua Bengio, and Jian Tang. 2021.
\newblock \href {https://openreview.net/forum?id=tGZu6DlbreV} {Rnnlogic: Learning logic rules for reasoning on knowledge graphs}.
\newblock In \emph{9th International Conference on Learning Representations, {ICLR} 2021, Virtual Event, Austria, May 3-7, 2021}. OpenReview.net.

\bibitem[{Sadeghian et~al.(2019)Sadeghian, Armandpour, Ding, and Wang}]{DBLP:conf/nips/SadeghianADW19}
Ali Sadeghian, Mohammadreza Armandpour, Patrick Ding, and Daisy~Zhe Wang. 2019.
\newblock {DRUM:} end-to-end differentiable rule mining on knowledge graphs.
\newblock In \emph{NeurIPS}, pages 15321--15331.

\bibitem[{Schick et~al.(2023)Schick, Dwivedi-Yu, Dessì, Raileanu, Lomeli, Zettlemoyer, Cancedda, and Scialom}]{schick_toolformer_2023}
Timo Schick, Jane Dwivedi-Yu, Roberto Dessì, Roberta Raileanu, Maria Lomeli, Luke Zettlemoyer, Nicola Cancedda, and Thomas Scialom. 2023.
\newblock \href {http://arxiv.org/abs/2302.04761} {Toolformer: {Language} {Models} {Can} {Teach} {Themselves} to {Use} {Tools}}.
\newblock \emph{arXiv preprint}.
\newblock ArXiv:2302.04761 [cs].

\bibitem[{Schlichtkrull et~al.(2018)Schlichtkrull, Kipf, Bloem, van~den Berg, Titov, and Welling}]{DBLP:conf/esws/SchlichtkrullKB18}
Michael~Sejr Schlichtkrull, Thomas~N. Kipf, Peter Bloem, Rianne van~den Berg, Ivan Titov, and Max Welling. 2018.
\newblock \href {https://doi.org/10.1007/978-3-319-93417-4\_38} {Modeling relational data with graph convolutional networks}.
\newblock In \emph{The Semantic Web - 15th International Conference, {ESWC} 2018, Heraklion, Crete, Greece, June 3-7, 2018, Proceedings}, volume 10843 of \emph{Lecture Notes in Computer Science}, pages 593--607. Springer.

\bibitem[{SHEN et~al.(2025)SHEN, Li, Meng, Cai, Qi, Zhang, Xu, and Ma}]{shen_shortcutsbench_2024_1}
Haiyang SHEN, Yue Li, Desong Meng, Dongqi Cai, Sheng Qi, Li~Zhang, Mengwei Xu, and Yun Ma. 2025.
\newblock \href {https://openreview.net/forum?id=kKILfPkhSz} {Shortcutsbench: A large-scale real-world benchmark for {API}-based agents}.
\newblock In \emph{The Thirteenth International Conference on Learning Representations}.

\bibitem[{Shen et~al.(2023)Shen, Song, Tan, Li, Lu, and Zhuang}]{shen_hugginggpt_2023_1}
Yongliang Shen, Kaitao Song, Xu~Tan, Dongsheng Li, Weiming Lu, and Yueting Zhuang. 2023.
\newblock \href {https://proceedings.neurips.cc/paper_files/paper/2023/file/77c33e6a367922d003ff102ffb92b658-Paper-Conference.pdf} {{HuggingGPT}: Solving {AI} tasks with {ChatGPT} and its friends in {Hugging Face}}.
\newblock In \emph{Advances in Neural Information Processing Systems}.

\bibitem[{Shen et~al.(2024)Shen, Song, Tan, Zhang, Ren, Yuan, Lu, Li, and Zhuang}]{shen_taskbench_2023_1}
Yongliang Shen, Kaitao Song, Xu~Tan, Wenqi Zhang, Kan Ren, Siyu Yuan, Weiming Lu, Dongsheng Li, and Yueting Zhuang. 2024.
\newblock \href {https://arxiv.org/abs/2311.18760} {{TaskBench}: Benchmarking large language models for task automation}.
\newblock In \emph{Advances in Neural Information Processing Systems}.

\bibitem[{Song et~al.(2023)Song, Xiong, Zhu, Wu, Qian, Song, Huang, Li, Wang, Yao, Tian, and Li}]{song_restgpt_2023}
Yifan Song, Weimin Xiong, Dawei Zhu, Wenhao Wu, Han Qian, Mingbo Song, Hailiang Huang, Cheng Li, Ke~Wang, Rong Yao, Ye~Tian, and Sujian Li. 2023.
\newblock \href {http://arxiv.org/abs/2306.06624} {{RestGPT}: {Connecting} {Large} {Language} {Models} with {Real}-{World} {RESTful} {APIs}}.
\newblock \emph{arXiv preprint}.
\newblock ArXiv:2306.06624 [cs].

\bibitem[{Sturua et~al.(2024)Sturua, Mohr, Akram, G{\"u}nther, Wang, Krimmel, Wang, Mastrapas, Koukounas, Wang et~al.}]{sturua2024jina}
Saba Sturua, Isabelle Mohr, Mohammad~Kalim Akram, Michael G{\"u}nther, Bo~Wang, Markus Krimmel, Feng Wang, Georgios Mastrapas, Andreas Koukounas, Nan Wang, and 1 others. 2024.
\newblock jina-embeddings-v3: Multilingual embeddings with task lora.
\newblock \emph{arXiv preprint arXiv:2409.10173}.

\bibitem[{Sui et~al.(2024)Sui, He, Ding, and Hooi}]{sui2024can}
Yuan Sui, Yufei He, Zifeng Ding, and Bryan Hooi. 2024.
\newblock Can knowledge graphs make large language models more trustworthy? an empirical study over open-ended question answering.
\newblock \emph{arXiv preprint arXiv:2410.08085}.

\bibitem[{Sun et~al.(2019)Sun, Deng, Nie, and Tang}]{DBLP:conf/iclr/SunDNT19}
Zhiqing Sun, Zhi{-}Hong Deng, Jian{-}Yun Nie, and Jian Tang. 2019.
\newblock \href {https://openreview.net/forum?id=HkgEQnRqYQ} {Rotate: Knowledge graph embedding by relational rotation in complex space}.
\newblock In \emph{7th International Conference on Learning Representations, {ICLR} 2019, New Orleans, LA, USA, May 6-9, 2019}. OpenReview.net.

\bibitem[{Talmor and Berant(2018)}]{talmor2018web}
Alon Talmor and Jonathan Berant. 2018.
\newblock The web as a knowledge-base for answering complex questions.
\newblock \emph{arXiv preprint arXiv:1803.06643}.

\bibitem[{Tang et~al.(2023)Tang, Deng, Lin, Han, Liang, Cao, and Sun}]{tang_toolalpaca_2023}
Qiaoyu Tang, Ziliang Deng, Hongyu Lin, Xianpei Han, Qiao Liang, Boxi Cao, and Le~Sun. 2023.
\newblock \href {http://arxiv.org/abs/2306.05301} {{ToolAlpaca}: {Generalized} {Tool} {Learning} for {Language} {Models} with 3000 {Simulated} {Cases}}.
\newblock \emph{arXiv preprint}.
\newblock ArXiv:2306.05301 [cs].

\bibitem[{Team(2025)}]{qwq32b}
Qwen Team. 2025.
\newblock \href {https://qwenlm.github.io/blog/qwq-32b/} {Qwq-32b: Embracing the power of reinforcement learning}.

\bibitem[{Teru et~al.(2020)Teru, Denis, and Hamilton}]{DBLP:conf/icml/TeruDH20}
Komal~K. Teru, Etienne~G. Denis, and William~L. Hamilton. 2020.
\newblock \href {http://proceedings.mlr.press/v119/teru20a.html} {Inductive relation prediction by subgraph reasoning}.
\newblock In \emph{Proceedings of the 37th International Conference on Machine Learning, {ICML} 2020, 13-18 July 2020, Virtual Event}, volume 119 of \emph{Proceedings of Machine Learning Research}, pages 9448--9457. {PMLR}.

\bibitem[{Toutanova et~al.(2015)Toutanova, Chen, Pantel, Poon, Choudhury, and Gamon}]{toutanova2015representing}
Kristina Toutanova, Danqi Chen, Patrick Pantel, Hoifung Poon, Pallavi Choudhury, and Michael Gamon. 2015.
\newblock Representing text for joint embedding of text and knowledge bases.
\newblock In \emph{EMNLP}.

\bibitem[{Trivedi et~al.(2022)Trivedi, Balasubramanian, Khot, and Sabharwal}]{trivedi2022musique}
Harsh Trivedi, Niranjan Balasubramanian, Tushar Khot, and Ashish Sabharwal. 2022.
\newblock Musique: Multihop questions via single-hop question composition.
\newblock \emph{Transactions of the Association for Computational Linguistics}, 10:539--554.

\bibitem[{Trouillon et~al.(2016)Trouillon, Welbl, Riedel, Gaussier, and Bouchard}]{TrouillonWRGB_2016}
Th{\'{e}}o Trouillon, Johannes Welbl, Sebastian Riedel, {\'{E}}ric Gaussier, and Guillaume Bouchard. 2016.
\newblock \href {http://proceedings.mlr.press/v48/trouillon16.html} {Complex embeddings for simple link prediction}.
\newblock In \emph{Proceedings of the 33nd International Conference on Machine Learning, {ICML} 2016, New York City, NY, USA, June 19-24, 2016}, volume~48 of \emph{{JMLR} Workshop and Conference Proceedings}, pages 2071--2080. JMLR.org.

\bibitem[{Wang et~al.(2024{\natexlab{a}})Wang, Luo, Chen, Mai, Guo, Dong, Xiaohua, Xuan, Li, Ma, and Gao}]{wang_mllm-tool_2024}
Chenyu Wang, Weixin Luo, Qianyu Chen, Haonan Mai, Jindi Guo, Sixun Dong, Xiaohua, Xuan, Zhengxin Li, Lin Ma, and Shenghua Gao. 2024{\natexlab{a}}.
\newblock \href {https://doi.org/10.48550/arXiv.2401.10727} {{MLLM}-{Tool}: {A} {Multimodal} {Large} {Language} {Model} {For} {Tool} {Agent} {Learning}}.
\newblock \emph{arXiv preprint}.
\newblock ArXiv:2401.10727 [cs].

\bibitem[{Wang et~al.(2024{\natexlab{b}})Wang, Wang, Xue, Xia, Cao, Liu, Pan, and Wong}]{wang_appbench_2024_1}
Hongru Wang, Rui Wang, Boyang Xue, Heming Xia, Jingtao Cao, Zeming Liu, Jeff~Z. Pan, and Kam-Fai Wong. 2024{\natexlab{b}}.
\newblock \href {https://doi.org/10.18653/v1/2024.emnlp-main.856} {{A}pp{B}ench: Planning of multiple {API}s from various {APP}s for complex user instruction}.
\newblock In \emph{Proceedings of the 2024 Conference on Empirical Methods in Natural Language Processing}, pages 15322--15336, Miami, Florida, USA. Association for Computational Linguistics.

\bibitem[{Wang et~al.(2024{\natexlab{c}})Wang, Wu, Wang, Liu, Song, Peng, Deng, Zhang, Wang, Peng, Zhang, Guo, Zhang, Su, and Zheng}]{wang_mtu-bench_2024}
Pei Wang, Yanan Wu, Zekun Wang, Jiaheng Liu, Xiaoshuai Song, Zhongyuan Peng, Ken Deng, Chenchen Zhang, Jiakai Wang, Junran Peng, Ge~Zhang, Hangyu Guo, Zhaoxiang Zhang, Wenbo Su, and Bo~Zheng. 2024{\natexlab{c}}.
\newblock \href {http://arxiv.org/abs/2410.11710} {{MTU}-{Bench}: {A} {Multi}-granularity {Tool}-{Use} {Benchmark} for {Large} {Language} {Models}}.
\newblock \emph{arXiv preprint}.
\newblock ArXiv:2410.11710.

\bibitem[{Wang et~al.(2024{\natexlab{d}})Wang, Shi, Wang, Lee, Yuan, Huang, and Lyu}]{wang_learning_2024}
Wenxuan Wang, Juluan Shi, Chaozheng Wang, Cheryl Lee, Youliang Yuan, Jen-tse Huang, and Michael~R. Lyu. 2024{\natexlab{d}}.
\newblock \href {https://doi.org/10.48550/arXiv.2409.00557} {Learning to {Ask}: {When} {LLMs} {Meet} {Unclear} {Instruction}}.
\newblock \emph{arXiv preprint}.
\newblock ArXiv:2409.00557 [cs].

\bibitem[{Wu et~al.(2024)Wu, Zhu, Han, Tan, Zhang, and Chen}]{wu_seal-tools_2024_1}
Mengsong Wu, Tong Zhu, Han Han, Chuanyuan Tan, Xiang Zhang, and Wenliang Chen. 2024.
\newblock \href {https://doi.org/10.1007/978-981-97-9434-8_29} {{Seal-Tools}: Self-instruct tool learning dataset for agent tuning and detailed benchmark}.
\newblock In \emph{Natural Language Processing and Chinese Computing: NLPCC 2024}, volume 15360 of \emph{Lecture Notes in Computer Science}, pages 372--384. Springer.

\bibitem[{Xiong et~al.(2017)Xiong, Hoang, and Wang}]{deeppath}
Wenhan Xiong, Thien Hoang, and William~Yang Wang. 2017.
\newblock Deeppath: A reinforcement learning method for knowledge graph reasoning.
\newblock In \emph{EMNLP}.

\bibitem[{Xu et~al.(2023)Xu, Hong, Li, Hu, Chen, and Zhang}]{xu_tool_2023}
Qiantong Xu, Fenglu Hong, Bo~Li, Changran Hu, Zhengyu Chen, and Jian Zhang. 2023.
\newblock \href {http://arxiv.org/abs/2305.16504} {On the {Tool} {Manipulation} {Capability} of {Open}-source {Large} {Language} {Models}}.
\newblock \emph{arXiv preprint}.
\newblock ArXiv:2305.16504 [cs].

\bibitem[{Yan et~al.(2022)Yan, Ma, Gao, Tang, and Chen}]{DBLP:conf/icml/YanMGT022}
Zuoyu Yan, Tengfei Ma, Liangcai Gao, Zhi Tang, and Chao Chen. 2022.
\newblock Cycle representation learning for inductive relation prediction.
\newblock In \emph{{ICML}}, volume 162 of \emph{Proceedings of Machine Learning Research}, pages 24895--24910. {PMLR}.

\bibitem[{Yang et~al.(2025)Yang, Li, Yang, Zhang, Hui, Zheng, Yu, Gao, Huang, Lv, Zheng, Liu, Zhou, Huang, Hu, Ge, Wei, Lin, Tang, Yang, Tu, Zhang, Yang, Yang, Zhou, Zhou, Lin, Dang, Bao, Yang, Yu, Deng, Li, Xue, Li, Zhang, Wang, Zhu, Men, Gao, Liu, Luo, Li, Tang, Yin, Ren, Wang, Zhang, Ren, Fan, Su, Zhang, Zhang, Wan, Liu, Wang, Cui, Zhang, Zhou, and Qiu}]{qwen3}
An~Yang, Anfeng Li, Baosong Yang, Beichen Zhang, Binyuan Hui, Bo~Zheng, Bowen Yu, Chang Gao, Chengen Huang, Chenxu Lv, Chujie Zheng, Dayiheng Liu, Fan Zhou, Fei Huang, Feng Hu, Hao Ge, Haoran Wei, Huan Lin, Jialong Tang, and 41 others. 2025.
\newblock Qwen3 technical report.
\newblock \emph{arXiv preprint arXiv:2505.09388}.

\bibitem[{Yang et~al.(2024)Yang, Yang, Hui, Zheng, Yu, Zhou, Li, Li, Liu, Huang, Dong, Wei, Lin, Tang, Wang, Yang, Tu, Zhang, Ma, Xu, Zhou, Bai, He, Lin, Dang, Lu, Chen, Yang, Li, Xue, Ni, Zhang, Wang, Peng, Men, Gao, Lin, Wang, Bai, Tan, Zhu, Li, Liu, Ge, Deng, Zhou, Ren, Zhang, Wei, Ren, Fan, Yao, Zhang, Wan, Chu, Liu, Cui, Zhang, and Fan}]{qwen2}
An~Yang, Baosong Yang, Binyuan Hui, Bo~Zheng, Bowen Yu, Chang Zhou, Chengpeng Li, Chengyuan Li, Dayiheng Liu, Fei Huang, Guanting Dong, Haoran Wei, Huan Lin, Jialong Tang, Jialin Wang, Jian Yang, Jianhong Tu, Jianwei Zhang, Jianxin Ma, and 40 others. 2024.
\newblock Qwen2 technical report.
\newblock \emph{arXiv preprint arXiv:2407.10671}.

\bibitem[{Yang et~al.(2017)Yang, Yang, and Cohen}]{DBLP:conf/nips/YangYC17}
Fan Yang, Zhilin Yang, and William~W. Cohen. 2017.
\newblock Differentiable learning of logical rules for knowledge base reasoning.
\newblock In \emph{{NIPS}}, pages 2319--2328.

\bibitem[{Yang et~al.(2023)Yang, Song, Li, Zhao, Ge, Li, and Shan}]{yang_gpt4tools_2023_1}
Rui Yang, Lin Song, Yanwei Li, Sijie Zhao, Yixiao Ge, Xiu Li, and Ying Shan. 2023.
\newblock \href {https://proceedings.neurips.cc/paper_files/paper/2023/file/e393677793767624f2821cec8bdd02f1-Paper-Conference.pdf} {{GPT4Tools}: Teaching large language model to use tools via self-instruction}.
\newblock In \emph{Advances in Neural Information Processing Systems}.

\bibitem[{Yang et~al.(2018)Yang, Qi, Zhang, Bengio, Cohen, Salakhutdinov, and Manning}]{yang2018hotpotqa}
Zhilin Yang, Peng Qi, Saizheng Zhang, Yoshua Bengio, William~W Cohen, Ruslan Salakhutdinov, and Christopher~D Manning. 2018.
\newblock Hotpotqa: A dataset for diverse, explainable multi-hop question answering.
\newblock \emph{arXiv preprint arXiv:1809.09600}.

\bibitem[{Ye et~al.(2025)Ye, Li, Gao, Huang, Wu, Li, Fan, Dou, Ji, Zhang, Gui, and Huang}]{ye_tooleyes_2025_1}
Junjie Ye, Guanyu Li, Songyang Gao, Caishuang Huang, Yilong Wu, Sixian Li, Xiaoran Fan, Shihan Dou, Tao Ji, Qi~Zhang, Tao Gui, and Xuanjing Huang. 2025.
\newblock \href {https://aclanthology.org/2025.coling-main.12/} {{ToolEyes}: Fine-grained evaluation for tool learning capabilities of large language models in real-world scenarios}.
\newblock In \emph{Proceedings of the 31st International Conference on Computational Linguistics}, pages 156--187, Abu Dhabi, UAE. Association for Computational Linguistics.

\bibitem[{Ye et~al.(2024{\natexlab{a}})Ye, Li, Li, Huang, Gao, Wu, Zhang, Gui, and Huang}]{ye_toolsword_2024_1}
Junjie Ye, Sixian Li, Guanyu Li, Caishuang Huang, Songyang Gao, Yilong Wu, Qi~Zhang, Tao Gui, and Xuanjing Huang. 2024{\natexlab{a}}.
\newblock \href {https://doi.org/10.18653/v1/2024.acl-long.119} {{T}ool{S}word: Unveiling safety issues of large language models in tool learning across three stages}.
\newblock In \emph{Proceedings of the 62nd Annual Meeting of the Association for Computational Linguistics (Volume 1: Long Papers)}, pages 2181--2211, Bangkok, Thailand. Association for Computational Linguistics.

\bibitem[{Ye et~al.(2024{\natexlab{b}})Ye, Wu, Gao, Huang, Li, Li, Fan, Zhang, Gui, and Huang}]{ye_rotbench_2024_1}
Junjie Ye, Yilong Wu, Songyang Gao, Caishuang Huang, Sixian Li, Guanyu Li, Xiaoran Fan, Qi~Zhang, Tao Gui, and Xuanjing Huang. 2024{\natexlab{b}}.
\newblock \href {https://doi.org/10.18653/v1/2024.emnlp-main.19} {{R}o{TB}ench: A multi-level benchmark for evaluating the robustness of large language models in tool learning}.
\newblock In \emph{Proceedings of the 2024 Conference on Empirical Methods in Natural Language Processing}, pages 313--333, Miami, Florida, USA. Association for Computational Linguistics.

\bibitem[{Yih et~al.(2016)Yih, Richardson, Meek, Chang, and Suh}]{yih2016value}
Wen-tau Yih, Matthew Richardson, Christopher Meek, Ming-Wei Chang, and Jina Suh. 2016.
\newblock The value of semantic parse labeling for knowledge base question answering.
\newblock In \emph{Proceedings of the 54th Annual Meeting of the Association for Computational Linguistics (Volume 2: Short Papers)}, pages 201--206.

\bibitem[{Yuan et~al.(2024)Yuan, Ning, Zhou, Yang, Li, Zhuang, Tan, Yao, Lin, Li et~al.}]{yuan2024lv}
Tao Yuan, Xuefei Ning, Dong Zhou, Zhijie Yang, Shiyao Li, Minghui Zhuang, Zheyue Tan, Zhuyu Yao, Dahua Lin, Boxun Li, and 1 others. 2024.
\newblock Lv-eval: A balanced long-context benchmark with 5 length levels up to 256k.
\newblock \emph{arXiv preprint arXiv:2402.05136}.

\bibitem[{Zhang and Yao(2022)}]{DBLP:conf/www/ZhangY22}
Yongqi Zhang and Quanming Yao. 2022.
\newblock Knowledge graph reasoning with relational digraph.
\newblock In \emph{{WWW}}, pages 912--924. {ACM}.

\bibitem[{Zhang et~al.(2018)Zhang, Dai, Kozareva, Smola, and Song}]{zhang2018variational}
Yuyu Zhang, Hanjun Dai, Zornitsa Kozareva, Alexander Smola, and Le~Song. 2018.
\newblock Variational reasoning for question answering with knowledge graph.
\newblock In \emph{Proceedings of the AAAI conference on artificial intelligence}, volume~32.

\bibitem[{ZHENG et~al.()ZHENG, WEI, LU, YIN, ZHANG, XU, CUI, SUN, CHEN, and QIU}]{YiningZHENG:0}
Yining ZHENG, Haiyang WEI, Jiahao LU, Linqi YIN, Yunke ZHANG, Chengguo XU, Hetao CUI, Tianxiang SUN, Shuang CHEN, and Xipeng QIU.
\newblock \href {https://doi.org/10.1007/s11704-025-41365-6} {Investigating effective llm-based in-context tool use: What matters and how to improve}.
\newblock \emph{Frontiers of Computer Science}, 0.

\bibitem[{Zhu et~al.(2021)Zhu, Zhang, Xhonneux, and Tang}]{DBLP:conf/nips/ZhuZXT21}
Zhaocheng Zhu, Zuobai Zhang, Louis{-}Pascal A.~C. Xhonneux, and Jian Tang. 2021.
\newblock \href {https://proceedings.neurips.cc/paper/2021/hash/f6a673f09493afcd8b129a0bcf1cd5bc-Abstract.html} {Neural bellman-ford networks: {A} general graph neural network framework for link prediction}.
\newblock In \emph{Advances in Neural Information Processing Systems 34: Annual Conference on Neural Information Processing Systems 2021, NeurIPS 2021, December 6-14, 2021, virtual}, pages 29476--29490.

\end{thebibliography}
\newpage
\appendix
\section{Additional Ablations}
\label{appendix:ablations_more}
In this section we show three important parts of our pipeline, checking how different extraction method will effect KG extraction and Tool use for starting. Next we show the KG extraction step is important by showing the degrading performancs of models with full KG. We also show the signal usage of KG will increase the Tool use performances. Last we show the simple pipeline for KG-based Tool-use is so atomic that we can use different LLMs in two steps respectively to get better performances. The ablations are done on FamilyTool-b if not mentioned specifically.

\subsection{Greedy Search vs. Relation Retrieval}
\label{appendix:ablation_gs_rr}
\begin{figure*}[htbp]
    \centering
    \includegraphics[width=\linewidth]{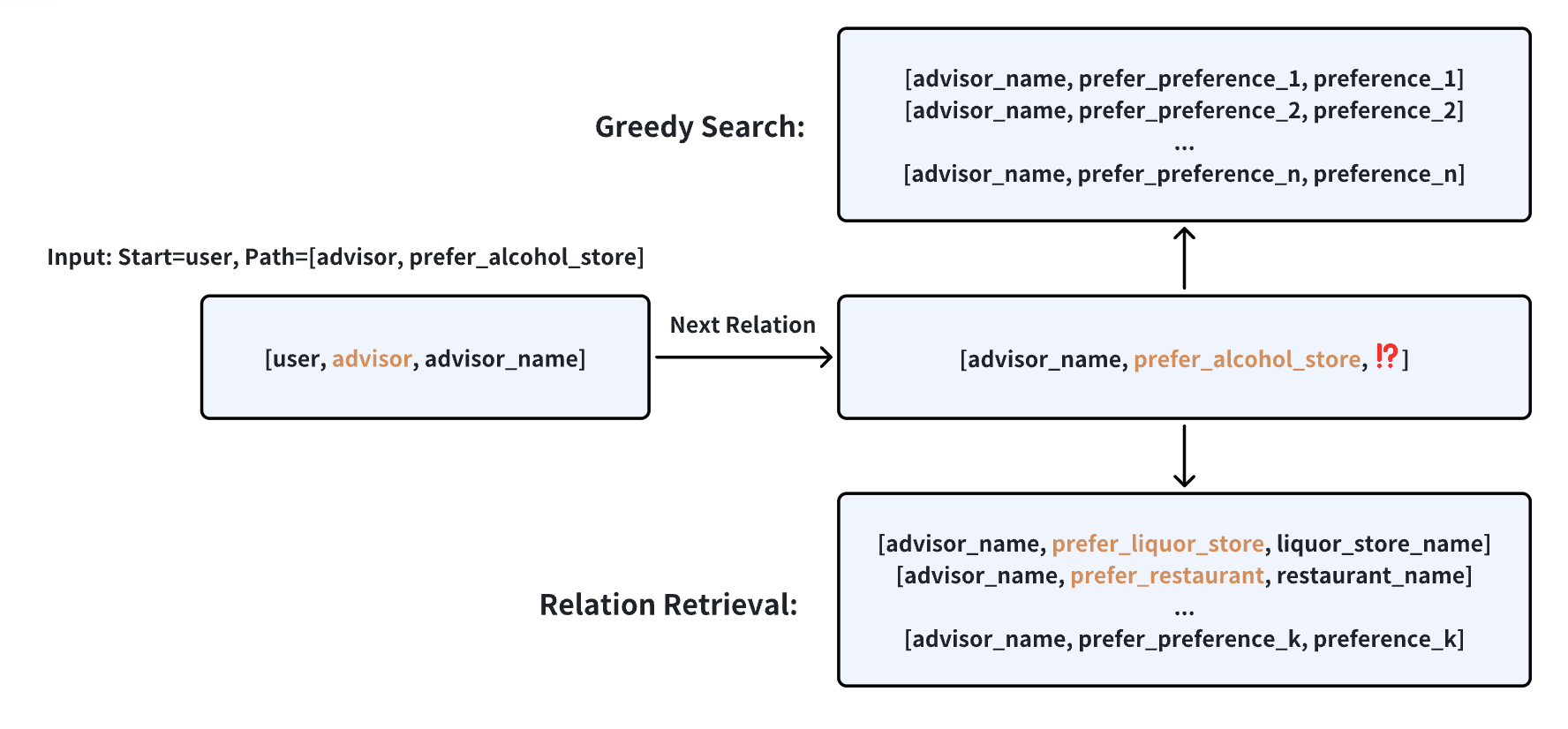}
    \caption{Path extraction example for demonstration of Greedy Search and Relation Retrieval.}
    \label{fig:pathextraction}
\end{figure*}

\subsubsection{KG Extraction}
\begin{table*}[htbp]
    \centering

    \resizebox{\textwidth}{!}{
        \begin{tabular}{lcccccc}
            \toprule
            LLM for KG &Path Extraction& EM & F1 &No-Hallucination & Coverage& Format Error 
\\
            \midrule
            Llama3.1-8B&GS & 43.89 & 59.33 & 52.42 &67.91 & 1.66 
\\
            Llama3.1-8B&RR & 41.82 & 65.45 & 98.73 &56.52 & 2.07 
\\
            \midrule
            Qwen2.5-7B&GS  & 43.89 & 60.72 & 51.47 &60.87 & 1.45 
\\
            Qwen2.5-7B&RR  & 39.75 & 64.63 & 100.00 &49.90 & 2.69 
\\
            \midrule
            Qwen2.5-32B&GS  & 59.42 & 74.35 & 85.53 &68.53 &1.24 
\\
            Qwen2.5-32B&RR  & 53.42 & 73.92 & 99.58 &59.21 &2.48 
\\
            \midrule
            QwQ-32B&GS  & 59.21 & 74.24 & 87.10 &69.98 & 3.73 
\\
            QwQ-32B&RR  & 54.04 & 73.47 & 93.25 &61.70 & 4.97 
\\
            \midrule
             Qwen3-8B&GS & 61.28 & 76.96 & 82.53 &75.57 & 1.66 
\\
            Qwen3-8B&RR & 56.11 & 76.01 & 97.45 &65.84 & 2.69 
\\\midrule
 Qwen3-8B-nothink& GS & 57.56 & 70.87 & 63.10 & 68.32 &1.24 
\\
 Qwen3-8B-nothink& RR & 52.38 & 71.41 & 100.00 & 58.59 &2.69 
\\
            \midrule
            Qwen3-32B&GS & 58.39 & 75.90 &95.39 &71.84 & 1.24 
\\
            Qwen3-32B&RR & 53.83 & 74.87 &99.79 &62.53 &2.69 
\\
            \midrule
            Qwen3-32B-nothink&GS & 54.24 & 73.17 &84.31 &72.46 & 1.04 
\\
             Qwen3-32B-nothink&RR & 49.48 & 72.65 &99.37 &61.70 & 2.07 
\\
            \midrule
            GPT-4o&GS  & 61.90 & 76.82 &89.41 &73.08 & 2.28 
\\
            GPT-4o&RR  & 56.31 & 75.59 &100.00 & 62.11 &3.52 
\\
            \midrule
            DeepSeek-V3&GS  & 65.42 &78.90 &92.68 &75.78 & 1.04 
\\
            DeepSeek-V3&RR  & 59.83 & 77.99 &100.00 &65.63 & 2.69 \\
\midrule
            o3-mini&GS  & 57.35 & 74.83 &93.90 &71.64 & 4.97 
\\
            o3-mini&RR  & 51.55 & 73.83 &100.00 & 62.11 &6.21 
\\
            \midrule
            DeepSeek-r1&GS  & 65.01 &79.38 &93.31 &74.53 & 1.04 
\\
            DeepSeek-r1&RR  & 59.42 & 77.86 &100.00 &64.60 & 2.48 \\

            \bottomrule
        \end{tabular}
    }
    \caption{Results for LLM KG extraction using GS and RR.}
    \label{tab:appendix-results_gsrrkgextraction}
\end{table*}
Coming to the comparison of Greedy Search and Relation Retrieval, Relation Retrieval cannot get better results than Greedy Search. This is because in FamilyTool, the largest distance in KG is 2, so when meeting fake relations in the second step, the Greedy Search extraction will definitely get the right entity. So Relation Retrieval is more useful in KGs with large diameters. Also, the retriever can introduce errors in relation retrieval without training. So in experiments in the main paper, we use Greedy Search for KG extraction without specific comments.

\textbf{Thinking in Qwen3.} The Qwen3 model integrates thinking mode, so we also evaluate its performance in KG extraction. We found that without thinking, the performances of Qwen3 models will decrease, which is reasonable for multi-hop reasoning.

\subsubsection{Tool Use}
Even in KG extraction RR fails in the comparison with GS. But GS can include more wrong paths and may have bad effects on the tool use  step. So we examine their performances in the tool use step in Table \ref{tab:ablation_GSvsRR_tooluse}. We found that for GPT-4o and DeepSeek-V3, even though Relation Retrieval performs worse than Greedy Search in the coverage of KG extraction, less noise can make them performs better in tool use step. 
\begin{table}[htbp]
    \centering

    \resizebox{0.48\textwidth}{!}{
        \begin{tabular}{l|ccc|ccc}
            \toprule
                        & \multicolumn{3}{c}{Greedy Search}& \multicolumn{3}{c}{Relation Retrieval}
            \\
            \midrule
            LLM for Tool Use& EM &Tool Acc. & Value Acc.& EM &Tool Acc. & Value Acc.\\
            \midrule
            Llama3.1-8B & 16.36 & 54.66 & 23.89 & 16.40 &55.50 & 23.50 
\\     \midrule
            Qwen2.5-7B  & 28.78 & 78.88 & 38.19 & 27.10 &79.50 & 36.50 
\\
             Qwen2.5-32B  & 28.99 &67.08 & 38.28 & 26.90 &65.20 & 36.50 
\\
            QwQ-32B  & 37.06 & 79.30 &45.47 & 26.90 &65.20 & 36.50 
\\     \midrule
            Qwen3-8B  & 40.37& 80.33&49.23& 34.80 &77.20 & 43.40 
\\
       
            Qwen3-32B& 36.02 & 80.33&46.06 & 35.40 &77.00 & 44.30 
\\     \midrule
            GPT-4o  & 19.67 & 76.81 &23.64 & 32.70 &78.70 & 42.30 
\\
             DeepSeek-V3  & 30.43 & 64.60 &39.23 & 37.33 &64.60 & 36.94 \\
            \bottomrule
        \end{tabular}
           
    }
 \caption{Results for LLM tool use with Extracted sub-KG with Greedy Search and Relation Retrieval.}
    \label{tab:ablation_GSvsRR_tooluse}
\end{table}


            
            

\subsection{Full KG in LLMs}
\label{ablation:fullkginllms}
In this ablation we show when without KG extraction step and adding the whole KG with 150+ links (a very small one compared to those in real-world scenarios) in the LLM input, how the LLMs can use the extra information. As shown in Table \ref{tab:ablation_fullkg_tooluse}, full KG will hugely degrade the performances of relatively small LLMs like Llama3.1-8B and Qwen2.5-7B. It has a relatively small effect on 32B models and Qwen3 Series, showing the strong understanding ability of reasoning models. In the end, results show that with full-KG, LLMs can difficultly extract useful information from KG, even with this small KG. All models show worse results than the results with extracted KG in Table \ref{tab:results_pipeline_tooluse} except GPT-4o, which has problems identifying entities as valid parameters. So in application, it is important to first extract a sub-KG from the full KG.

\begin{table}[htbp]
    \centering

    \resizebox{0.45\textwidth}{!}{
        \begin{tabular}{lccc}
            \toprule
            LLM for Tool Use & EM &Tool Acc. & Value Acc.\\
            \midrule
            Llama3.1-8B & 2.69 & 30.64 & 6.63 
\\\midrule
            Qwen2.5-7B  & 7.45 & 69.98 & 17.86 
\\
            Qwen2.5-32B  & 21.74 & 64.18 &29.49 
\\
            QwQ-32B  & 33.13 & 82.40 &43.40 
\\\midrule
                
            Qwen3-8B  & 33.54 & 83.23 &43.43 
\\
             Qwen3-32B& 35.40 & 85.71 &46.21 
\\
            \midrule
            GPT-4o  & 20.08 & 75.98 &24.74 
\\
            DeepSeek-V3  & 24.22 & 69.57 &33.82 \\
            \bottomrule
        \end{tabular}
    }
    \caption{Results for LLM tool-use with Full KG.}
    \label{tab:ablation_fullkg_tooluse}
\end{table}

\subsection{True Values in KG}
\label{sec:ablation_true_values_inKG}
In this section, we use true values in the KG (eg. fried chicken instead of food$\_$0000) and evaluate models on FamilyTool-e with gold KG. Results are shown in Table \ref{tab:ablation_realkg_tooluse}. We found that all models performs worse compared to using signal in KG, showing the importance of using signals in KG. Also, we found that although Qwen2.5-32B still cannot outperform Qwen2.5-7B, the gap is smaller, which is the result of using the true values so that 32B model will know it's valid and not ask for a multi-round conversation. As a validation, the average tool call of Qwen2.5-32B increases from 78.68 to 84.40.
\begin{table}[htbp]
    \centering

    \resizebox{0.45\textwidth}{!}{
        \begin{tabular}{lccc}
            \toprule
            LLM for Tool Use & EM &Tool Acc. & Value Acc.\\
            \midrule
            Llama3.1-8B & 24.62 & 49.45 & 29.39 
\\\midrule
            Qwen2.5-7B  & 43.96 & 80.00 & 52.97 
\\
            Qwen2.5-32B  & 36.04 & 68.35 &44.11 
\\
            QwQ-32B  & 46.15 & 77.58 &54.41 
\\\midrule
                
            Qwen3-8B  & 45.71 & 78.46 &54.60 
\\
             Qwen3-32B& 49.67 & 79.34 &58.13 
\\
            \midrule
            GPT-4o  & 29.89 & 77.80 &33.99 
\\
            DeepSeek-V3  & 41.54 & 64.18 &49.06 \\
            \bottomrule
        \end{tabular}
    }
    \caption{Results for LLM tool-use with real KG values. }
    \label{tab:ablation_realkg_tooluse}
\end{table}

\subsection{Cross-LLMs}
In previous experiments we can find that some models can show better performances in the KG extraction step, which encourages us to see how different LLMs for KG extraction can effect the tool use. Will the same LLM achieve good results? We show the results in Table \ref{tab:ablation_cross_LLM}.

\label{ablation:cross_llms}
\begin{table*}
    \centering

\resizebox{\textwidth}{!}{\begin{tabular}{l|c|ccc|cc|cc}
        \toprule
        LLM-KG\text{||} Tool Use &  Llama3.1-8B &Qwen2.5-7B  & Qwen2.5-32B  & QwQ-32B & Qwen3-8B& Qwen3-32B& GPT-4o & DeepSeek-V3 
\\
        
        \midrule
        Llama3.1-8B &     16.36 & 29.40 & 28.16 & 31.26& 32.51 & 32.92 & 17.18 & 24.22 
\\
        
         Qwen2.5-7B  &     13.25 & 28.78 & 25.47 & 28.99 & 30.85 & 30.43 & 15.11 & 23.81 
\\ Qwen2.5-32B  &     18.01 & 34.37 & 28.99 & 34.58 & 34.78 & 34.58 & 17.81 & 28.16 
\\ QwQ-32B  &     17.60 & 34.99 & 32.30 & 37.06 & 37.27 & 39.13 & 18.01 & 28.36 
\\ Qwen3-8B  &     19.67 & 39.75 & \textbf{33.75} & 39.13 & 40.37 & 37.89 & 19.25 & 29.81 
\\ Qwen3-32B&     16.77 & 36.65 & 30.02 & 35.20 & 37.06 & 36.02 & 18.63 & 28.16 
\\ GPT-4o  &     \textbf{22.15} & 38.10 & 32.92 & 37.06 & 39.13 & 38.30 & 19.67 & 30.02 
\\ DeepSeek-V3  &     20.29 & \textbf{39.96} & 33.54 & \textbf{39.54} & \textbf{41.41} & 39.13 & \textbf{21.33} & 30.43 
\\ o3-mini&     20.08 & 36.65 & 31.68 & 35.61 & 38.10 & 38.51 & 19.88 & 30.23 
\\ DeepSeek-R1&     20.70 & 38.51 & 33.54 & 37.68 & 41.20 & \textbf{40.17} & 20.70 & \textbf{31.47} \\
        \bottomrule
    \end{tabular}}
    
    \caption{EM Results for LLM tool use with Extracted sub-KG. We bold the best performances in the column, showing the best LLM for KG extraction for different LLMs for tool use.}
    \label{tab:ablation_cross_LLM}
\end{table*}

With the results of KG extraction in Table \ref{tab:results_kgextraction}, we found that the higher in KG extraction, the higher in tool use . There is no matching magic for using the same LLMs in two steps. With the best Performances in KG extraction, using DeepSeek-V3 for KG extraction makes half of the models performs best in the tool use step.

\section{Prompt templates for FamilyTool Generation}
In this section we show the prompt templates we use for construction of FamilyTool.
\subsection{Prompt Template (Preference Synthesis)}
The Prompt template is "\textit{I need you to generate a knowledge graph(KG) based on the existing KG of 3 family members. The existing KG is formatting as : [[Alice, husband, Jack], [Jack, wife, Alice],[Alice, son, Bob],[Jack, son, Bob],[Bob, mother, Alice],[Bob, father, Jack]]. You can add the preferences of those people, like preferred movie, music, etc, to generate query for a give n API. Note: You only need to provide me with a JSON format question for your answer. The answer example is: $EXAMPLE$. The API document is $API\_DOC$.}", in which the example is:
\begin{tcolorbox}[colback=white!95!gray,colframe=gray!50!black,rounded corners, label={Example}, title={EXAMPLE}] 
\begin{lstlisting}[breaklines=true, xleftmargin=0pt, breakindent=0pt, columns=fullflexible] 
{
    "name": "order_pizza",
    "question": [
        "<speak>Speaker: Alice</speak> Can you order a pizza with my son's  preferred crust type and toppings? ([Alice, son, Bob], [Bob, prefer_pizza_crust, pizza_crust_0000], [Bob, prefer_pizza_topping, pizza_topping_0000])",
        "<speak>Speaker: Jack</speak> I'd like to get a pizza with the toppings my son likes most. ([Jack, son, Bob], [Bob, prefer_pizza_topping, pizza_topping_0000])",
            ]
        }.
\end{lstlisting} 
\end{tcolorbox}.

\subsection{Prompt Template (Query Synthesis)}

The Prompt template is "\textit{This is a text generation game; I will provide you with an API\ The 'description' field describes the functionality of this API\ The content in "parameters " is called a slot,  
    At the same time, I will provide you with a knowledge graph composed of triplets in the form of [car owner, child, son] as additional information about the car owner's family relationships. For example, when using a chain composed of [car owner, child, son] and [son, favorite supermarket, supermarket A], emphasize that ***** should not include the last entity * * * in the generated answer. Also, place the link information used in a specific format in the sentence, for example: call my child's favorite supermarket ([car owner, child, son], [son, favorite supermarket, supermarket A]), and place the used triplets at the end of the sentence. Be careful not to say the link in the sentence. The last entity of the last triplet, such as' Supermarket A 'here, please describe the last entity using the first two words of the triplet. 
    Generate sentences that call this API in the voice of passengers or drivers. Emphasize that the sentences you generate must conform to the driving environment and the normal way of speaking for people. The sentences should be appropriately colloquial
    The slot is our scoring point. If the questions you generate contain content that can fill the slot, you will receive a certain score. Our scoring rules are as follows:
    1. Each generated question earns one point. 
    2. Each question that contains content that can be filled into a slot earns 0.5 points. If the slot uses link information from the knowledge graph, an additional 5 points will be added
    3. Count the number of times all slots are covered, and add the number of times the slot with the least coverage is covered to the final score. Note: You only need to provide me with a JSON format question for your answer, you don't need to calculate the score. The answer example is: $EXAMPLE$. Please remember that the content that can be filled into slots in the question should be specific and use the link information from the knowledge graph**** Once again, it should be emphasized that the generated questions should be concise and in line with the way humans speak*** And the slot in the 'required' field of the API
    It must exist in the question. If your question is about contacts, then there should also be this part in the triplet link. Now, I will provide you with an API, please challenge yourself and get the highest possible score! Keep it up! $API\_DOC$. Next, I will provide you with a knowledge graph with additional information. $KG$}"

To generate different speakers, we assume a car scenerio that all passengers and driver can talk to the car agent in the prompt template.

\subsection{Prompt Template (Answer Synthesis)}

The Prompt template is "\textit{This is a slot replenishment game; I will provide you with an API\ The 'description' field describes the functionality of this API\ The content in \"parameters \" is called a slot. This is the API I gave you:$API\_DOC$. Please help me find the content related to the slot in the question I gave you. My answer example is: $EXAMPLE$. Here are the questions and data format I want to give you. $data$. Output format considerations: 1. The 'parameters' in the answer should include the filling result of the API, which should be the last entity of the last triplet. 2. The parameters' that can be filled should be repeated at the end. 3. If you want to say anything extra, just give me an answer in sample format, which should be a JSON type data. 4. Generate corresponding answers for all questions, don't miss them, and all content should be in English"}"

\subsection{Prompt Template (Query Check)}

The Prompt template is "\textit{You are a function call artificial intelligence. Now, your task is to determine whether the user is requesting a service from you. You need to return the reasoning and the result to me. Wrap them in <thought></thought> and <answer></answer> respectively. If you believe your assistance is needed, the answer should be <answer>Yes</answer>, otherwise <answer>No</answer>. The <thought> section should be no more than 50 words in length.}"

\subsection{Prompt Template (Answer Check)}
The Prompt template is "\textit{"You are a function call artificial intelligence. However, you have encountered an issue: some of the function call data is incomplete. Your task now is to check whether the data is valid.You will be given the user's request, detailed documentation for the tool, and the invocation statement. There may be errors within.You should carefully consider how the tool is used, whether the invocation conforms to the specifications, and whether it meets the user's request.You need to explain your reasoning and your conclusion. Wrap them in <thought></thought> and <answer></answer> respectively.If you believe the tool invocation is structurally incorrect, the result should be <answer>Misformatted</answer>.If the format is correct but it doesn't match the user's intent, the result should be <answer>Misaligned</answer>.If the format is correct and matches the user's intent, but the tool's capabilities are limited and fail to satisfy the user, the result should be <answer>Limited</answer>.If the invocation matches the user's intent but contains informal parameter issues (e.g., arbitrarily setting boolean parameters, including actions in a value), the result should be <answer>Informal</answer>.If the invocation is well-formatted, aligns with the user's intent, and follows conventions, the result should be <answer>Proper</answer>. It is recommended to only use the above terms in the <answer> section. Use a new term only if none of the above apply. The <thought> section should be no longer than 50 words.}"

\subsection{Prompt Template (Dataset Extension)}

The Prompt template is "\textit{This is a task of forging data. I now have a piece of data about an API call, but now my KG has expanded. What you need to do is to transform the original data and replace the person in the question based on the API call data I gave you and the new KG I gave you. The new data must meet the following requirements:
The new data must use a KG of 3 hops or more, that is, the number of triples in the brackets must be three or more, for example:
1$EXAMPLE_1$.
2$EXAMPLE_2$.
3$EXAMPLE_3$.
hint: Don't talk about grandmother and grandfather in general. Mother's mother is maternal grandmother, mother's father is maternal grandfather, father's father is paternal grandfather, father's mother is paternal grandmother. Only when the speaker is Bob, there will be the situation of grandfather and grandmother!!
hint: When the speaker is Alice, you can replace it with my husband's mother/father, my son's teacher. When the speaker is Jack, you can replace it with my wife's mother/father, my son's teacher. When the speaker is Bob, you can replace it with father/mother's father/mother.
hint: The triple must contain one of ["William", "Charles", "Margaret", "Elizabeth", "Mr. Smith"]
You only need to give me an answer in json format, the format is consistent with the original data
The original data is here: $ORIG\_DATA$, new kg: $KG\_NEW$ }", in which $KG\_NEW$ is the links of new KG, excluding those in $\mathcal{G}_b$.

\section{More Experimental Details}
\label{appendix:implementationdetails}
\subsection{Models}
All LLMs are using the temperature of 0.7 for sampling. The performances of LLMs are relatively stable so we only run once for each experiment. Inference is applied by vLLM~\cite{kwon2023efficient}.

\begin{table*}[htbp]
\centering
\footnotesize
\begin{tabular}{|l|c|c|c|c|c|c|}
\hline
\textbf{Models} & \textbf{\# Para} & \textbf{Launch Time} & \textbf{Max Tokens} & \textbf{Scaling} & \textbf{Corporation} & \textbf{License} \\
\hline
GPT-4o & / & May 13, 2024 & 128,000 & Effort & OpenAI & Proprietary \\
\hline
o3-mini & / & Jan 31, 2025 & 200,000 & Effort & OpenAI & Proprietary \\
\hline
DeepSeek-v3-0324 & 671B & Mar 25, 2025 & 131,072 & Budget & DeepSeek & Open Source \\
\hline
DeepSeek-r1 & 671B & Mar 25, 2025 & 131,072 & Budget & DeepSeek & Open Source \\
\hline
Llama-3.1-8B & 8B & Jul 23, 2024 & 131,072 & Budget & Meta & Llama License \\
\hline
Qwen-2.5-7B & 7B & Sep 19, 2024 & 131,072 & Budget & Alibaba & Apache 2.0 \\
\hline
Qwen-2.5-32B & 32B & Sep 19, 2024 & 131,072 & Budget & Alibaba & Apache 2.0 \\
\hline
QwQ-32B & 32B & March 6, 2025 & 131,072 & Budget & Alibaba & Apache 2.0 \\
\hline
Qwen3-8B & 8B & Apr 29, 2025 & 131,072 & Budget & Alibaba & Qwen License \\
\hline
Qwen3-32B & 32B & Apr 29, 2025 & 131,072 & Budget & Alibaba & Qwen License \\
\hline
\end{tabular}
\caption{Large language models evaluated in our experiments with specifications and characteristics.}
\label{tables:model-detail}
\end{table*}

\subsection{Metrics}

For KG extraction, we use EM, F1, No-Hallucination, Coverage and Format Error to evaluate the results. For a single sample, the extracted link set is $\mathcal{M}$ and the golden link set si $\mathcal{G}$. If they are the same, then EM is 1. And F1 is calculated as below like that in multi-class classification.
\begin{align}
    &\text{TP} = |\mathcal{G} \cap \mathcal{M}|,  \\
    &\text{FP} = |\mathcal{M} \setminus \mathcal{G}|, \\
    &\text{FN} = |\mathcal{G} \setminus \mathcal{M}| ,\\
     &\text{Precision} = \frac{\text{TP}}{\text{TP} + \text{FP}}, \\
    &\text{Recall} = \frac{\text{TP}}{\text{TP} + \text{FN}}, \\
     &\text{F1} = 2 \times \frac{\text{Precision} \times \text{Recall}}{\text{Precision} + \text{Recall}}.
\end{align}
Coverage is to show whether the used KG entity of tool is covered in the sub-KG from KG extraction. No-Hallucination shows how many relation paths in KG extractionhave no fake relations. For tool use, we use EM, Tool Acc., Value Acc.. EM shows if the functional calling is exactly true. Tool Acc. shows if the model can choose right tools from candidate tools. Value Acc. computes the accuracy of treating each parameter of a tool as a test sample separately. For results demonstration, we show the average value of the metrics across test set.

\subsection{KG Relations}
We enclose the KG in the supplemental materials. The relations include relationships, preferences of travel, appointment, hotel, dining, transport, service, multimedias and miscellaneous ones.

\subsection{Hard Query Examples with 4 And 6 Hops}

\begin{tcolorbox}[colback=white!95!gray,colframe=gray!50!black,rounded corners, label={Example4hops}, title={EXAMPLE with 4 hops}] 
\begin{lstlisting}[breaklines=true, xleftmargin=0pt, breakindent=0pt, columns=fullflexible] 
{
    "name": "SearchTrain",
    "question": 
        "<speak>Speaker: Bob</speak> Search for a train leaving Dad's favorite spot arriving at Mom's favorite location, provide full details.  The extra information for the query is ([Bob, father, Jack], [Jack, prefer_dining_location, dining_location_0001], [Bob, mother, Alice], [Alice, prefer_dining_location, dining_location_0002])."
        }.
\end{lstlisting} 

\end{tcolorbox}

\begin{tcolorbox}[colback=white!95!gray,colframe=gray!50!black,rounded corners, label={Example6hops}, title={EXAMPLE with 6 hops}] 
\begin{lstlisting}[breaklines=true, xleftmargin=0pt, breakindent=0pt, columns=fullflexible] 
{
    "name": "SearchFlights",
    "question": 
        "<speak>Speaker: Bob</speak> Can you check flights between my paternal father's favorite city and my paternal grandmother's preferred destination?   The extra information for the query is ([Bob, father, Jack], [Jack, father, William], [William, prefer_travel_city, travel_city_0007], [Bob, father, Jack], [Jack, mother, Elizabeth], [Elizabeth, prefer_travel_to, travel_to_0004])."
        }.
\end{lstlisting} 

\end{tcolorbox}


\end{document}